\documentclass{article}

\usepackage[final]{neurips_2024}

\usepackage[utf8]{inputenc} 
\usepackage[T1]{fontenc}    
\usepackage{hyperref}       
\usepackage{url}            
\usepackage{booktabs}       
\usepackage{amsfonts}       
\usepackage{nicefrac}       
\usepackage{microtype}      
\usepackage{xcolor}         
\usepackage{amsmath,graphicx}
\usepackage{amsmath,amssymb,,subcaption}
\usepackage{amsthm}
\usepackage{mathtools}
\usepackage{dsfont}
\newtheorem{proposition}{Proposition} 
\newtheorem{assumption}{Assumption}  
  
\newtheorem{lemma}{Lemma}

\newtheorem{theorem}{Theorem}

\DeclareMathOperator*{\argmin}{arg\,min}

\newcommand\norm[1]{\left\lVert#1\right\rVert}
\title{Localized Adaptive Risk Control}

\author{%
Matteo Zecchin \quad Osvaldo Simeone \\
  Centre for Intelligent Information Processing Systems\\
  Department of Engineering\\
  King’s College London\\
  London, United Kingdom \\
  \texttt{\{matteo.1.zecchin,osvaldo.simeone\}@kcl.ac.uk} \\
}

\begin{document}
	
	\maketitle
    \begin{abstract}

        Adaptive Risk Control (ARC) is an online calibration strategy based on set prediction that offers worst-case deterministic long-term risk control, as well as statistical marginal coverage guarantees. ARC adjusts the size of the prediction set by varying a single scalar threshold based on feedback from past decisions. In this work, we introduce Localized Adaptive Risk Control (L-ARC), an online calibration scheme that targets statistical localized risk guarantees ranging from conditional risk to marginal risk, while preserving the worst-case performance of ARC. L-ARC updates a threshold function within a reproducing kernel Hilbert space (RKHS), with the kernel determining the level of localization of the statistical risk guarantee. The theoretical results highlight a trade-off between localization of the statistical risk and convergence speed to the long-term risk target. Thanks to localization, L-ARC is demonstrated via experiments to produce prediction sets with risk guarantees across different data subpopulations, significantly improving the fairness of the calibrated model for tasks such as image segmentation and beam selection in wireless networks.
    \end{abstract}
    \section{Introduction}

    Adaptive risk control (ARC), also known as online risk control, is a powerful tool for reliable decision-making in online settings where feedback is obtained after each decision \citep{gibbs2021adaptive,feldman2022achieving}. ARC finds applications in domains, such as finance, robotics, and health, in which it is important to ensure reliability in forecasting, optimization, or control of complex systems \citep{wisniewski2020application,lekeufack2023conformal,zhang2023bayesian,zecchin2024forking}. While providing worst-case deterministic guarantees of reliability, ARC may distribute such guarantees \emph{unevenly} in the input space, favoring a subpopulation of inputs at the detriment of another subpopulation. 
    
\begin{figure}
        \centering
        \includegraphics[width=0.9\textwidth]{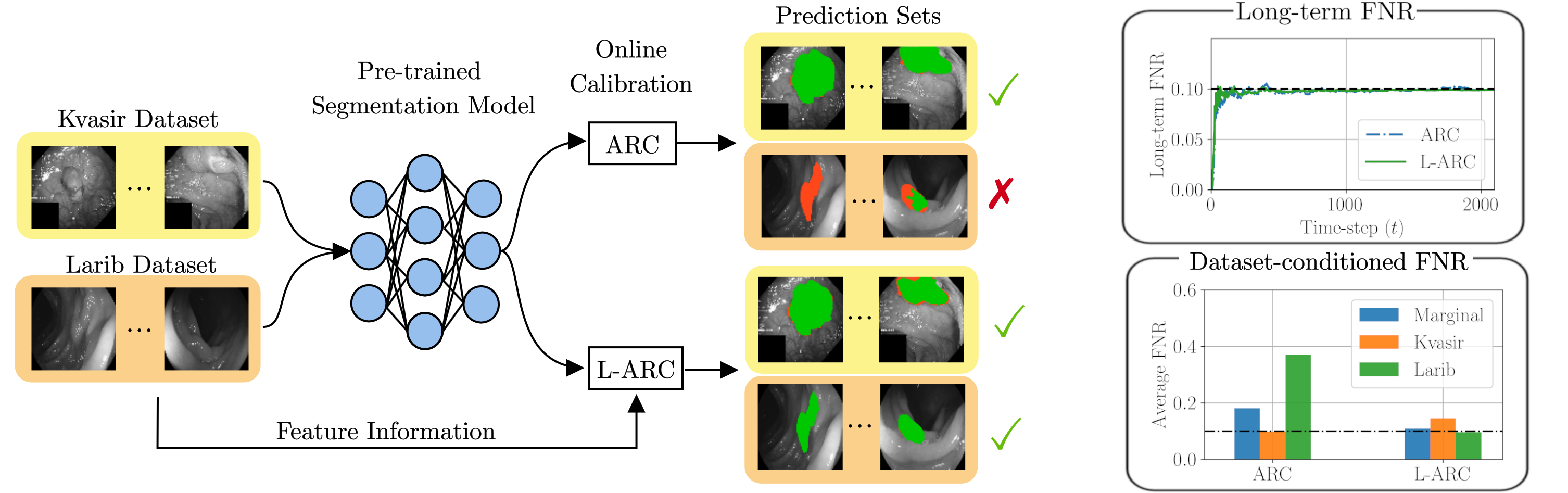}
        \caption{Calibration of a tumor segmentation model via ARC \citep{angelopoulos2024conformal} and the proposed localized ARC, L-ARC. Calibration data comprises images from multiple sources, namely, the Kvasir data set \citep{jha2020kvasir} and the ETIS-LaribPolypDB data set \citep{silva2014toward}. Both ARC and L-ARC achieve worst-case deterministic long-term risk control in terms of false negative rate (FNR). However, ARC does so by prioritizing Kvasir samples at the detriment of the Larib data source, for which the model has poor FNR performance. In contrast, L-ARC can yield uniformly satisfactory performance for both data subpopulations.}
     \label{fig:example}
    \end{figure}

    As an example, consider the tumor segmentation task illustrated in Figure \ref{fig:example}. In this setting, the objective is to calibrate a pre-trained segmentation model to generate masks that accurately identify tumor areas according to a user-defined reliability level \citep{yu2016integrating}. The calibration process typically involves combining data from various datasets, such as those collected from different hospitals. For an online setting, as visualized in the figure, ARC achieves the desired long-term reliability in terms of false negative ratio. However, it does so by prioritizing certain datasets, resulting in unsatisfactory performance on other data sources. Such behavior is particularly dangerous, as it may result in some subpopulations being poorly diagnosed. This paper addresses this shortcoming of ARC by proposing a novel \emph{localized} variant of ARC.
    \subsection{Adaptive Risk Control}
    To elaborate, consider an online decision-making scenario in which inputs are provided sequentially to a pre-trained model.  At each time step $t\geq 1$, the model observes a feature vector $X_t$, and based on a bounded \emph{non-conformity scoring function} $s:\mathcal{X}\times \mathcal{Y}\to [0,S_{\textrm{max}}]$ and a threshold $\lambda_t\in\mathbb{R}$, it outputs a prediction set 
    \begin{align}
        C_t=C(X_t,\lambda_t)=\{y\in\mathcal{Y}: s(X_t,y)\leq \lambda_t\},
        \label{eq:aci_set}
    \end{align}
    where $\mathcal{Y}$ is the domain of the target variable $Y$. 
    After each time step $t$, the model receives feedback in the form of a loss function
    \begin{align}
        L_t=\mathcal{L}(C_t,Y_t)
        \label{eq:loss_t}
    \end{align}
    that is assumed to be non-negative, upper bounded by $B<\infty$ and non-increasing in the predicted set size $|C_t|$. A notable example is the miscoverage loss
    \begin{align}
        \mathcal{L}(C,y)=\mathds{1}\{y\notin C\}.
        \label{eq:cov_loss}
    \end{align} 
    Accordingly, for an input-output sequence $\{(X_t,Y_t)\}^T_{t=1}$ the performance of the set predictions $\{C_t\}^T_{t=1}$ in (\ref{eq:aci_set}) can be gauged via the cumulative risk
    \begin{align}
    \label{eq:cum_risk}
    \bar{L}(T)=\frac{1}{T}\sum^T_{t=1} \mathcal{L}(C_t,Y_t)=\frac{1}{T}\sum^T_{t=1} L_t.
    \end{align}
    
    For a user-specified loss level $\alpha $ and a learning rate sequence $\{\eta_t\}^T_{t=1}$, ARC updates the threshold $\lambda_t$ in (\ref{eq:aci_set}) as  \citep{feldman2022achieving}
    \begin{align}
        \lambda_{t+1}=\lambda_t+\eta_t(L_t-\alpha ),
        \label{eq:aci_update_rule}
    \end{align}
    where $L_t-\alpha $ measures the discrepancy between the current loss (\ref{eq:loss_t}) and the target $\alpha $.
    For step size decreasing as $\eta_t=\eta_1 t^{-1/2}$ for $a\in(0,1)$ and an arbitrary $\eta_1>0$, the results in \citep{angelopoulos2024online} imply that the update rule (\ref{eq:aci_update_rule}) guarantees that the cumulative risk (\ref{eq:cum_risk}) {\color{black} for the miscoverage loss (\ref{eq:cov_loss})} converges to target level $\alpha $ for \emph{any} data sequence $\{(X_t,Y_t)\}_{t\geq 1}$ as
    \begin{align}
        \left|\bar{L}(T)-\alpha \right|\leq \frac{S_{\textrm{max}}+\eta_1 B}{\sqrt{T}},
        \label{eq:guarantee_aci}
    \end{align}
    thus offering a worst-case deterministic long-term guarantee.
    Furthermore when data are generated i.i.d. as $(X_t,Y_t)\sim P_{XY}$ for all $t\geq 1$, in the special case of the miscoverage loss (\ref{eq:cov_loss}), the set predictor produced by (\ref{eq:aci_update_rule}) enjoys the asymptotic \emph{marginal} coverage guarantee
    \begin{align}
        \lim_{T\to\infty} \Pr\left[Y\notin C_T\right]\stackrel{p}{=} \alpha  ,
        \label{eq:coverage_aci}
    \end{align}
      where the probability is computed with respect to the test sample $(X,Y)\sim P_{XY}$, which is independent of the sequence of samples $\{(X_t,Y_t)\}^T_{t=1}$, and the convergence is in probability with respect to the sequence $\{(X_t,Y_t)\}_{t\geq 1}$. Note that in \citep{angelopoulos2024online}, a stronger version of (\ref{eq:coverage_aci}) is provided, in which the limit holds almost surely. 
    \subsection{Conditional and Localized Risk}
    The convergence guarantee (\ref{eq:coverage_aci}) for ARC is marginalized over the covariate $X$. Therefore, there is no guarantee that the conditional miscoverage $\Pr\left[Y\notin C_T|X=x\right]$ is smaller than the target $\alpha $. This problem is particularly relevant for high-stakes applications in which it is important to ensure a homogeneous level of reliability across different regions of the input space, such as across subpopulations. That said, even when the set predictor $C(X|\mathcal{D}_{\textrm{cal}})$ is obtained based on an offline calibration data set $\mathcal{D}_{\textrm{cal}}$ with i.i.d. data $(X,Y)\sim P_{XY}$, it is generally impossible to control the conditional miscoverage probability as
    \begin{align}
        \Pr\left[Y\notin C(X|\mathcal{D}_{\textrm{cal}})|X=x\right]\leq \alpha \, \text{ for all } x\in \mathcal{X}
        \label{eq:cond_cov}
    \end{align}
    without making further assumptions about the distribution $P_{XY}$ or producing uninformative prediction sets \citep{vovk2012conditional,foygel2021limits}.
    
\begin{figure}
        \centering
        \includegraphics[width=\textwidth]{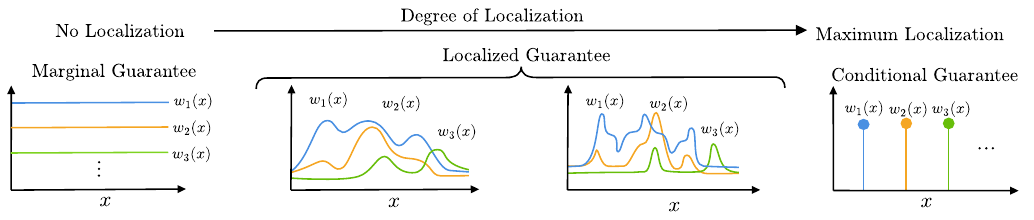}
        \caption{The degree of localization in L-ARC is dictated by the choice of the reweighting function class $\mathcal{W}$ via the marginal-to-conditional guarantee (\ref{eq:relaxation}). At the leftmost extreme, we illustrate constant reweighting functions, for which marginal guarantees are recovered. At the rightmost extreme, reweighting with maximal localization given by Dirac delta functions for which the criterion (\ref{eq:relaxation}) corresponds to a conditional guarantee. In between the two extremes lie function sets $\mathcal{W}$ with an intermediate level of localization yielding localized guarantees.}
        \label{fig:localization}
    \end{figure}

    A relaxed marginal-to-conditional guarantee was considered by \cite{gibbs2023conformal}, which relaxed the marginal miscoverage requirement (\ref{eq:cond_cov}) as
    \begin{align}
    \label{eq:relaxation}
         \mathbb{E}_{X,Y,\mathcal{D}_{\textrm{cal}}}\left[\frac{w(X)}{\mathbb{E}_X[w(X)]}\mathds{1}\{Y \notin C(X|\mathcal{D}_{\textrm{cal}})\}\right]\leq \alpha  \text{ for all } w(\cdot)\in\mathcal{W},
    \end{align}
    where $\mathcal{W}$ is a set of non-negative reweighting functions, and the expectation is taken over the joint distribution of the calibration data $\mathcal{D}_{\textrm{cal}}$ and the test pair $(X,Y)$. Note that with a singleton set $\mathcal{W}$ encompassing a single constant function, e.g., $w(x)=1$, the criterion (\ref{eq:relaxation}) reduces to marginal coverage. Furthermore, as illustrated in Figure \ref{fig:localization}, depending on the degree of localization of the functions in set $\mathcal{W}$, the criterion (\ref{eq:relaxation}) interpolates between marginal and conditional guarantees.
    
    At the one extreme, a marginal guarantee like (\ref{eq:coverage_aci}) is recovered when the reweighting functions are constant. Conversely, at the other extreme, conditional guarantees as in (\ref{eq:cond_cov}) emerge when the reweighting functions are maximally localized, i.e., when $\mathcal{W}=\{w(x)=\delta(x-\mu): \mu\in\mathcal{X}\}$, where $\delta(x)$ denotes the Dirac delta function. In between these two extremes, one obtains an intermediate degree of localization.  For example, this can be done by considering reweighting functions such as
    \begin{align}
        \label{eq:example_weightings}
        \mathcal{W}=\left\{w(x)\hspace{-0.2em}=\hspace{-0.2em}\sum^\infty_{i=1}\beta_i\left(\kappa\exp\left(-\frac{\norm{x-\mu_i}^2}{l}\right)+1\right)\hspace{-0.2em}: \mathbb{E}_X[w(X)]>0, \text{ and }  w(x)\geq 0 \ \forall x\in\mathcal{X} \right\},
    \end{align}
   where $l\geq 0$ is a fixed length scale, $\kappa\geq 0$ is a fixed scaling parameter, and $\norm{\cdot}$ denotes the Euclidean norm. Furthermore, function $w(x)$ may also depend on the output of the pre-trained model, supporting calibration requirements via constraints of the form (\ref{eq:relaxation}) \citep{zhang2024fair}.
    
    In \cite{gibbs2023conformal}, the authors demonstrated that it is possible to design \emph{offline} set predictors $C(X|\mathcal{D}_{\textrm{cal}})$ that \emph{approximately} control risk (\ref{eq:relaxation}), with an approximation gap that depends on the degree of localization of the family $\mathcal{W}$ of weighting functions.
    \subsection{Localized Risk Control}
    \label{sec:informal_localized_risk_control}
   Motivated by the importance of conditional risk guarantees, we propose Localized ARC (L-ARC), a novel online calibration algorithm that produces prediction sets with localized statistical risk control guarantees as in (\ref{eq:relaxation}), while also retaining the worst-case deterministic long-term guarantees (\ref{eq:guarantee_aci}) of ARC. Unlike \cite{gibbs2023conformal}, our work focuses on \emph{online} settings in which calibration is carried out sequentially based on feedback received on past decisions.       

    The key technical innovation of L-ARC lies in the way set predictions are constructed. As detailed in Section \ref{sec:LARC},  L-ARC prediction sets replace the single threshold in (\ref{eq:aci_set}) with a threshold function $g(\cdot)$ mapping covariate $X$ to a localized threshold value $g(X)$. The threshold function is adapted in an online fashion within a reproducing kernel Hilbert space (RKHS) family $\mathcal{G}$ based on an input data stream and loss feedback. The choice of the RKHS family determines the family $\mathcal{W}$ of weighting functions in the statistical guarantee of the form (\ref{eq:relaxation}), thus dictating the desired level of localization.
    
    The main technical results, presented in Section \ref{sec:rel_guarantees}, are as follows.

    \begin{itemize}
        \item In the case of i.i.d. sequences, $(X_t,Y_t)\sim P_{XY}$ for all $t\geq 1$, L-ARC provides localized statistical risk guarantees where the reweighting class $\mathcal{W}$ corresponds to all non-negative functions $w\in\mathcal{G}$ with a positive mean under distribution $P_{XY}$. More precisely, given a target loss value $\alpha $, the time-averaged threshold function 
        \begin{align}
        \label{eq:time_averaged}
        \bar{g}_T(\cdot)=\frac{1}{T}\sum^{T}_{t=1}g_t(\cdot),
        \end{align}
        ensures that for any function $w\in \mathcal{W}$,  {\color{black} the limit
    	\begin{align}
        \label{eq:informal_stat_rel}
    		\limsup_{T\to\infty}\mathbb{E}_{X,Y}\left[\frac{w(X)}{\mathbb{E}_X[w(X)]}\mathcal{L}(C(X,\bar{g}_T),Y)\right] \stackrel{p}{\leq}  \alpha +A(\mathcal{G},w)
    	\end{align}
        holds, where convergence is in probability with respect to the sequence $\{(X_t,Y_t)\}_{t\geq 1}$ and the average is over the test pair $(X,Y)$. The gap $A(\mathcal{G},w)$ depends on both the RKHS $\mathcal{G}$ and function $w$; it increases with the level of localization of the functions in the RKHS $\mathcal{G}$; and it equals zero in the case of constant threshold functions, recovering (\ref{eq:coverage_aci}) for the special case of the miscoverage loss.}
        
        \item Furthermore, for an arbitrary sequence $\{(X_t,Y_t)\}_{t\geq1}$ L-ARC has a cumulative loss that converges to a neighborhood of the nominal reliability level $\alpha $ as
    	\begin{align}
            \label{eq:informal_det_rel}
    		\left|\frac{1}{T}\sum^{T}_{t=1}\mathcal{L}(C(X_t,g_t),Y_t)-\alpha \right|\leq  \frac{B(\mathcal{G})}{\sqrt{T}}+C(\mathcal{G}),
    	\end{align}
    	where $B(\mathcal{G})$ and $C(\mathcal{G})$ are terms that increase with the level of localization of the function in the RKHS $\mathcal{G}$. The quantity $C(\mathcal{G})$ equals zero in the case of constant threshold functions, recovering the guarantee (\ref{eq:guarantee_aci}) of ARC.
    \end{itemize} 
    
    In Section \ref{sec:exp} we showcase the superior conditional risk control properties of L-ARC as compared to ARC for the task of electricity demand forecasting, tumor segmentation, and beam selection in wireless networks.

	\section{Localized Adaptive Risk Control} 
    \label{sec:LARC}
    \subsection{Setting}
    Unlike the ARC prediction set (\ref{eq:aci_set}), L-ARC adopts prediction sets that are defined based on a threshold function $g_t:\mathcal{X}\to\mathbb{R}$. Specifically, at each time $t\geq 1$ the L-ARC prediction set is obtained based on a non-conformity scoring function $s:\mathcal{X}\times \mathcal{Y}\to \mathbb{R}$ as
    \begin{align}
		C_t=C(X_t,g_t):=\left\{y\in\mathcal{Y}:s(X_t,y)\leq g_t(X_t)\right\}.
        \label{eq:LARC_set}
	\end{align} 
    By (\ref{eq:LARC_set}), the threshold $g_t(X_t)$ is localized, i.e., it is selected as a function of the current input $X_t$. In this paper, we consider threshold functions of the form 
     \begin{align}
        \label{eq:LARC_threshold}
         g_t(\cdot)=f_t(\cdot)+c_t,
     \end{align}
     where $c_t\in\mathbb{R}$ is a constant and function $f_t(\cdot)$ belongs to a reproducing kernel Hilbert space (RKHS) $\mathcal{H}$ associated to a kernel $k(\cdot,\cdot):\mathcal{X}\times \mathcal{X}\to \mathbb{R}$ with inner product $\langle\cdot,\cdot\rangle_{\mathcal{H}}$ and norm $\norm{\cdot}_\mathcal{H}$.  Note that the threshold function $g_t(\cdot)$ belongs to the RKHS $\mathcal{G}$ determined by the kernel $k'(\cdot,\cdot)=k(\cdot,\cdot)+1$.

     We focus on the online learning setting, in which at every time set $t\geq 1$, the model observes an input feature $X_t$, produces a set $C_t$, and receives as feedback the loss $L_t=\mathcal{L}(C_t,Y_t)$. Note that label $Y_t$ may not be directly observed, and only the loss $\mathcal{L}(C_t,Y_t)$ may be recorded. Based on the observed sequence of features $X_t$ and feedback $L_t$, we are interested in producing prediction sets as in (\ref{eq:LARC_set}) that satisfy the reliability guarantees (\ref{eq:informal_stat_rel}) and (\ref{eq:informal_det_rel}), with a reweighting function set $\mathcal{W}$ encompassing all non-negative functions $w(\cdot)\in\mathcal{G}$ with a positive mean $\mathbb{E}_X[w(X)]$ under distribution $P_{X}$, i.e.,
      \begin{align}
      \label{eq:covariate_shifts_set}
        \mathcal{W}=\{w(\cdot)\in\mathcal{G}: \  \ \mathbb{E}_X[w(X)]>0, \text{ and }  w(x)\geq 0 \text{ for all } x\in\mathcal{X}\}.
    \end{align} 
     Importantly, as detailed below, the level of localization in guarantee (\ref{eq:informal_stat_rel}) depends on the choice of the kernel $k(\cdot,\cdot)$. 
     
    \subsection{L-ARC}
    Given a regularization parameter $\lambda>0$ and a learning rate $\eta_t\leq 1/\lambda$, L-ARC updates the threshold function $g_t(\cdot)=f_t(\cdot)+c_t$ in (\ref{eq:LARC_set}) based on the recursive formulas
	\begin{align}
		c_{t+1}&=c_{t}-\eta_t(\alpha -L_t)\label{eq:constant_component}\\
		f_{t+1}(\cdot)&=(1-\lambda\eta_t)f_t(\cdot)-\eta_t(\alpha -L_t)k(X_t,\cdot)\label{eq:kernel_update},
	\end{align}
    with $f_1(\cdot)=0$ and $c_1=0$. In order to implement the update (\ref{eq:constant_component})-(\ref{eq:kernel_update}), it is useful to rewrite the function $g_{t+1}(\cdot)$ as
	\begin{align}
		g_{t+1}(\cdot)=\sum^{t}_{i=1}a^i_{t+1} k(X_i,\cdot)+c_{t+1},
        \label{eq:threshold}
	\end{align}
	where the coefficients $\{a^i_{t+1}\}^t_{i=1}$ are recursively defined as 
	\begin{align}
		a^t_{t+1}&=-\eta_t(\alpha -L_t)\label{eq:coeffsLARC_1}\\
		a^i_{t+1}&=(1-\eta_t\lambda)a^i_{t}, \quad \text{ for }i=1,2,\dots,t-1.
        \label{eq:coeffsLARC}
	\end{align}	
    Accordingly, if the loss $L_t$ is larger than the long-term target $\alpha $, the update rule (\ref{eq:coeffsLARC_1})-(\ref{eq:coeffsLARC}) increases the function $g_{t+1}(\cdot)$ around the current input $X_t$, while decreasing it around the previous inputs $X_1,\dots,X_{t-1}$. Intuitively, this change enhances the reliability for inputs in the neighborhood of $X_t$.

    It is important to note that, at any time $t$, computing the threshold function (\ref{eq:threshold}) requires storing the coefficients $\{a^i_{t}\}^{t-1}_{i=1}$ and $c_{t}$, as well as the input data $\{X_t\}^t_{i=1}$. Consequently, L-ARC has a linear memory requirement in $t$, which is a known limitation of non-parametric learning in online settings \citep{koppel2020optimally}. Previous research has explored methods that trade memory efficiency for accuracy \citep{kivinen2004online}. In Appendix \ref{sec:mem_eff}, we build on these approaches to present a memory-efficient variant of L-ARC that allows for a trade-off between localized risk control and memory requirements. 
    
    \subsection{Theoretical Guarantees}
    \label{sec:rel_guarantees}
    In this section, we formalize the theoretical guarantees of L-ARC, which were informally stated in Section \ref{sec:informal_localized_risk_control} as (\ref{eq:informal_stat_rel}) and (\ref{eq:informal_det_rel}).
	\begin{assumption}[Stationary and bounded kernel]
		\label{ass:lower_bound_k}
		The kernel function is stationary, i.e., $k(x,x')=\tilde{k}(\norm{x-x'})$, for some non-negative function $\tilde{k}(\cdot)$, which is $\rho$-Lipschitz for some $\rho>0$, upper bounded by $\kappa< \infty$, and coercive, i.e., $\lim_{z\to \infty}\tilde{k}(z)=0$.
	\end{assumption}
    Many well-known stationary kernels, such as the radial basis function (RBF), Cauchy, and triangular kernels, satisfy Assumption \ref{ass:lower_bound_k}.  
  The smoothness parameter $\rho$ and the maximum value of the kernel function $\kappa$ determine the localization of the threshold function $g_t(\cdot)\in\mathcal{G}$.  For example, the set of functions $\mathcal{W}$ defined in (\ref{eq:example_weightings}) corresponds to the function class (\ref{eq:covariate_shifts_set}) associated with the RKHS defined by the raised RBF kernel $k(x,x')=\kappa\exp(-\norm{x-x'}^2/l)+1$, with length scale $l=2e(\kappa/\rho)^2 $. As illustrated in Figure \ref{fig:localization}, by increasing $\kappa$ and $\rho$, we obtain functions with an increasing level of localization, ranging from constant functions to maximally localized functions. 
    \begin{assumption}[Bounded non-conformity scores]
		\label{ass:nc_score}
		The non-conformity scoring function is non-negative and bounded, i.e., $s(x,y)\leq S_{\textrm{max}}<\infty$ for any pair $(x,y)\in\mathcal{X}\times\mathcal{Y}$. 
	\end{assumption}   

	\begin{assumption}[Bounded and monotone loss]
		\label{ass:loss_bounded}
		The loss function is non-negative; bounded, i.e., $\mathcal{L}(C,Y)\leq B < \infty$  for any $C\subseteq{\mathcal{Y}}$ and $Y\in\mathcal{Y}$; and monotonic, in the sense that for prediction sets $C'$ and $C$ such that $C'\subseteq C$, the inequality $\mathcal{L}(C,Y)\leq\mathcal{L}(C',Y)$ holds for any $Y\in\mathcal{Y}$.
	\end{assumption}

	\subsubsection{Statistical Localized Risk Control}
    \label{sec:iid}
     To prove the localized statistical guarantee (\ref{eq:informal_stat_rel}) we will make the following assumption.
    {\color{black}
    \begin{assumption}[Strictly decreasing loss]
		\label{ass:unique_minimizer}
	    For any fixed threshold function $g(\cdot)\in\mathcal{G}$, the loss  $\mathbb{E}_{Y}[\mathcal{L}(C(X,g),Y)|X=x]$ is strictly decreasing in the threshold $g(x)$ for any $x\in\mathcal{X}$.
 	\end{assumption}
    \begin{assumption}[Left-continuous loss]
		\label{ass:left_continuity}
		For any fixed threshold function $g(\cdot)\in\mathcal{G}$, the loss  $\mathcal{L}(C(x,g+h),y)$ is left-continuous in $h\in\mathbb{R}$ for any $(x,y)\in \mathcal{X}\times\mathcal{Y}$.
 	\end{assumption}}

	\begin{theorem}
    \label{th:iid}
        Fix a user-defined target reliability $\alpha $. For any regularization parameter $\lambda>0$ and any learning rate sequence $\eta_t=\eta_1 t^{-1/2}<1/\lambda$, for some $\eta_1>0$, given a sequence $\{(X_t,Y_t)\}^T_{t=1}$ of i.i.d. samples from $P_{XY}$, the time-averaged threshold function (\ref{eq:time_averaged}) satisfies the limit
		\begin{align}
            \label{eq:formal_stat_guarantee}
			\limsup_{T\to\infty}	\mathbb{E}_{X,Y}\left[\frac{w(X)}{\mathbb{E}_X[w(X)]}\mathcal{L}(C(X,\bar{g}_T),Y)\right] \stackrel{p}{\leq} \alpha +\kappa B\frac{\norm{f_w}_{\mathcal{H}}}{{\mathbb{E}_X[w(X)]}},
		\end{align}
		for any weighting function $w(\cdot)=f_w(\cdot)+c_w\in\mathcal{W}$ where the expectation is with respect to the test sample $(X,Y)$.
	\end{theorem}
	\begin{proof}
    See Appendix \ref{app:th2}.
	\end{proof}
    {\color{black}By (\ref{eq:formal_stat_guarantee}), the average localized loss converges in probability to a quantity that can be bounded by the target $\alpha $ with a gap $A(\mathcal{G},w)$ that increases with the level of localization $\kappa$.}

	\subsubsection{Worst-Case Deterministic Long-Term Risk Control}
    \label{sec:longrun}
	\begin{theorem}
		\label{th1}
        Fix a user-defined target reliability $\alpha $. For any regularization parameter $\lambda>0$ and any learning rate sequence $\eta_t=\eta_1 t^{-1/2}< 1/\lambda$ with $\eta_1>0$, given any sequence $\{(X_t,Y_t)\}^T_{t=1}$ with bounded input $\norm{X_t}\leq D<\infty$, L-ARC produces a sequence of threshold functions $\{g_t(\cdot)\}^T_{t=1}$ in (\ref{eq:threshold}) that satisfy the inequality
        \begin{align}
        \label{eq:mistake_bound_th}
			\left|\frac{1}{T}\sum^T_{t=1} \mathcal{L}(C(X_t,g_t),Y_t)-\alpha \right|\leq   \frac{1}{\sqrt{T}}\left(\frac{S_{\textrm{max}}}{\eta_0 }+\frac{4B \sqrt{\rho{\color{black}\kappa D}}}{\eta_0\lambda}+2B (2\kappa+1)\right) +{\color{black}\kappa B}.
		\end{align}
	\end{theorem}
	\begin{proof}
		We defer the proof to Appendix \ref{app:th1}.
	\end{proof}
	Formalizing the upper bound in (\ref{eq:informal_det_rel}), Theorem \ref{th1} states that the difference between the long-term cumulative risk and the target reliability level $\alpha $ decreases with a rate $B(\mathcal{G})T^{-1/2}$ to a value $C(\mathcal{G})=\kappa B$ that is increasing with the maximum value of the kernel $\kappa$. In the special case, $\kappa=0$ which corresponds to no localization, the right-hand side of (\ref{eq:mistake_bound_th}) vanishes in $T$, recovering ARC long-term guarantee (\ref{eq:guarantee_aci}). 
	
	\section{Experiments} 
	\label{sec:exp}
     In this section, we explore the worst-case long-term and statistical localized risk control performance of L-ARC as compared to ARC. Firstly, we address the task of electricity demand forecasting, utilizing data from the Elec2 dataset \citep{harries1999splice}. Next, we present an experiment focusing on tumor segmentation, where the data comprises i.i.d. samples drawn from various image datasets \citep{jha2020kvasir,bernal2015wm,bernal2012towards,silva2014toward,vazquez2017benchmark}. Finally, we study a problem in the domain of communication engineering by focusing on beam selection, a key task in wireless systems \citep{ali2017millimeter}. A further example concerning applications with calibration constraints can be found in Appendix \ref{app:exp_condidence}. Unless stated otherwise, we instantiate L-ARC with the RBF kernel $k(x,x')=\kappa \exp(-\norm{x-x'}^2/l)$ with $\kappa=1$, length scale $l=1$ and regularization parameter $\lambda=10^{-4}$. With a smaller length scale $l$, we obtain increasingly localized weighting functions.  All the experiments are conducted on a consumer-grade Mac Mini with an M1 chip. The simulation code is available at  \url{https://github.com/kclip/localized-adaptive-risk-control.git}.
    
\subsection{Electricity Demand}
    
    \begin{figure}
    \centering
    \begin{subfigure}{.5\textwidth}
      \centering
      \includegraphics[width=0.9\linewidth]{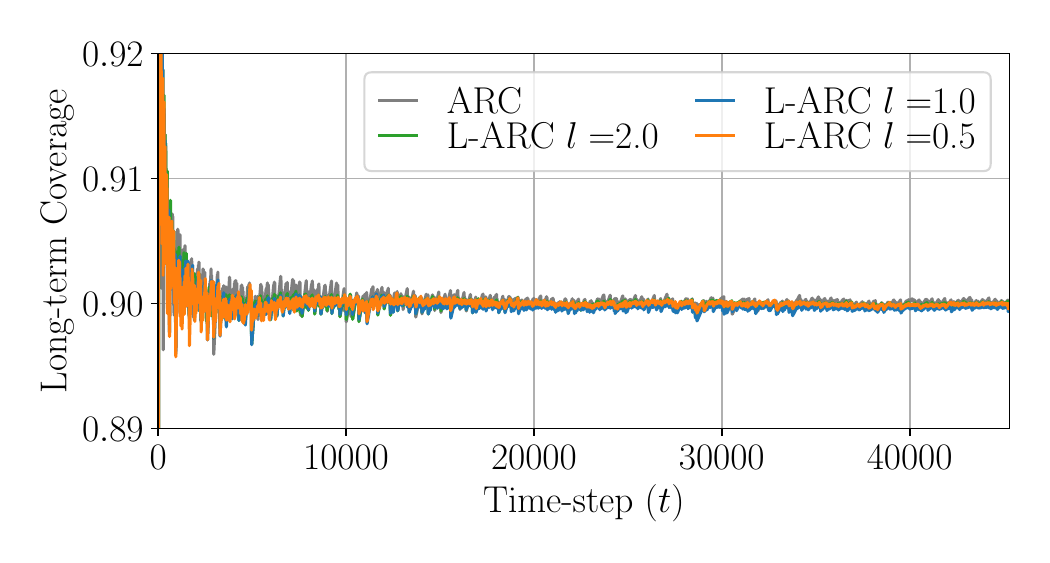}
    \end{subfigure}%
    \begin{subfigure}{.5\textwidth}
      \centering
      \includegraphics[width=0.9\linewidth]{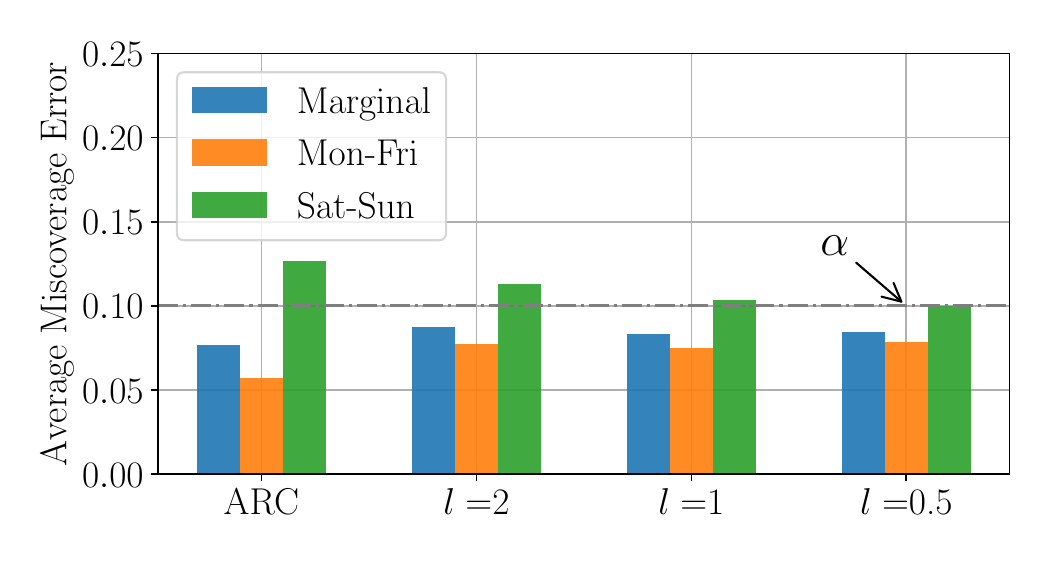}
    \end{subfigure}
    \caption{Long-term coverage (left) and average miscoverage error (right), marginalized and conditioned on weekdays and weekends. for ARC and L-ARC with varying values of the localization parameter $l$ on the Elec2 dataset.}
    \label{fig:elec}
    \end{figure}
    The Elec2 dataset comprises $T=45312$ hourly recordings of electricity demands in New South Wales, Australia. The data sequence $\{Y_{t}\}^T_{t=1}$ is subject to distribution shifts due to fluctuations in demand over time, such as between day and night or between weekdays and weekends. We adopt a setup akin to that of \cite{angelopoulos2024online}, wherein the even-time data samples are used for online calibration while odd-time data samples are used to evaluate coverage after calibration. At time $t$, the observed covariate $X_t$ corresponds to the past time series $Y_{1:t-1}$, and the forecasted electricity demand $\hat{Y}_t$ is obtained based on a moving average computed from demand data collected within the preceding 24 to 48 hours. We produce prediction sets $C_t$ based on the non-conformity score $s(X_t,Y_t)=|\hat{Y}_t-Y_t|$ and we target a miscoverage rate $\alpha =0.1$ using the miscoverage loss (\ref{eq:cov_loss}). Both ARC and L-ARC use the learning rate $\eta_t=t^{-1/2}$.  L-ARC is instantiated with the RBF kernel $k(x,x')=\kappa \exp(-\norm{\phi(x)-\phi(x')}^2/l)$, where $\phi(x)$ is a 7-dimensional feature vector corresponding to the daily average electricity demand during the past 7 days.
    
    In the left panel of Figure \ref{fig:elec}, we report the cumulative miscoverage error of ARC and L-ARC for different values of the localization parameter $l$. All algorithms converge to the desired coverage level of 0.9 in the long-term. The right panel of Figure \ref{fig:elec}, displays the average miscoverage error on the hold-out dataset at convergence. We specifically evaluate both the marginalized miscoverage rate and the conditional miscoverage rate separately over weekdays and weekends.  L-ARC is shown to reduce the weekend coverage error rate as compared to ARC providing balanced coverage as the length scale $l$ decreases.
    
    \subsection{Tumor Image Segmentation}
     \label{sec:exp_tumore_seg}
     \begin{figure}
    \centering
    \begin{subfigure}{.32\textwidth}
      \centering
      \includegraphics[width=1\linewidth]{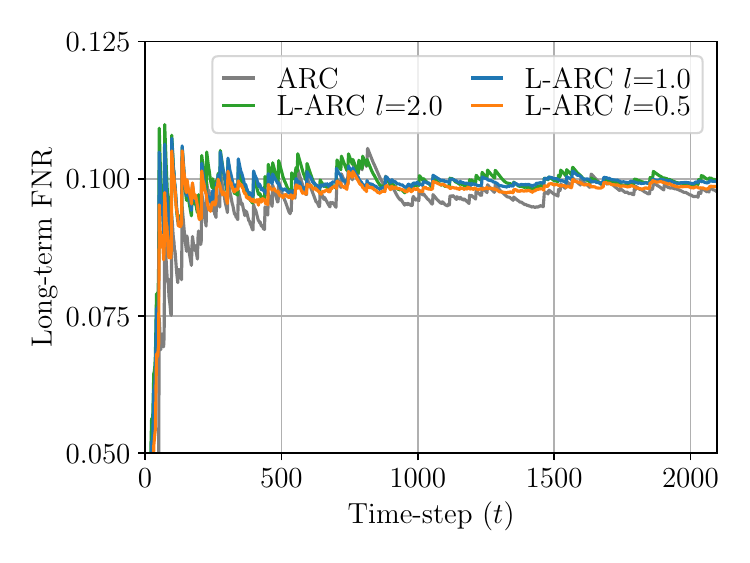}
    \end{subfigure}%
    \begin{subfigure}{.32\textwidth}
      \centering
      \includegraphics[width=1\linewidth]{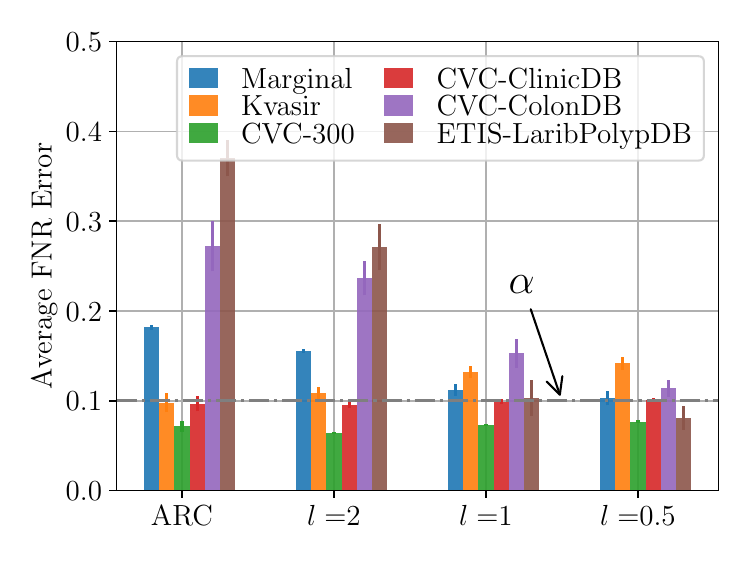}
    \end{subfigure}
\begin{subfigure}{.32\textwidth}
      \centering
      \includegraphics[width=1\linewidth]{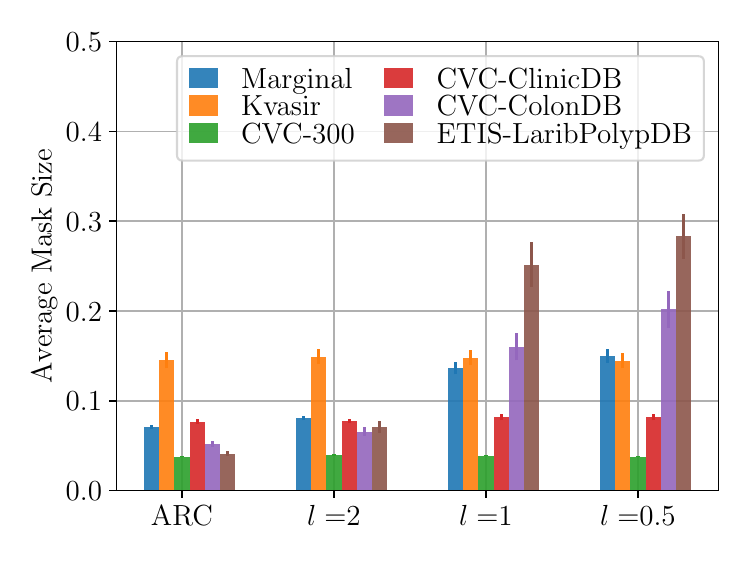}
    \end{subfigure}
    \caption{
    Long-term FNR (left), average FNR across different data sources (center), and average mask size across different data sources (right) for ARC and L-ARC with varying values of the localization parameter $l$ for the task of tumor segmentation \citep{fan2020pranet}. }
    \label{fig:polyp}
    \end{figure}

    In this section, we focus on the task of calibrating a predictive model for tumor segmentation. Here, the feature vector $X_t$ represents a $d_{\textrm{H}}\times d_{\textrm{W}}$ image, while the label $Y_t\subseteq\mathcal{P}$ identifies a subset of the image pixels $\mathcal{P}=\{(1,1),\dots,(d_{\textrm{H}},d_{\textrm{W}})\}$ that encompasses the tumor region. As in \cite{angelopoulos2022conformal}, the dataset is a compilation of samples from several open-source online repositories: Kvasir, CVC-300, CVC-ColonDB, CVC-ClinicDB, and ETIS-LaribDB. We reserve 50 samples from each repository for testing the performance post-calibration, while the remaining $T=2098$ samples are used for online calibration.
    Predicted sets are obtained by applying a threshold $g(X_t)$ to the pixel-wise logits $f(p_{\textrm{H}},p_{\textrm{W}})$ generated by the PraNet segmentation model \citep{fan2020pranet}, with the objective of controlling the false negative ratio (FNR) $\mathcal{L}(C_t,Y_t)=1-|C_t\cap Y_t|/|Y_t|$.
    Both ARC and L-ARC are run using the same decaying learning rate $\eta_t=0.1t^{-1/2}$.  L-ARC is instantiated with the RBF kernel $k(x,x')=\kappa \exp(-\norm{\phi(x)-\phi(x')}^2/l)$, where $\phi(x)$ is a 5-dimensional feature vector obtained via the principal component analysis (PCA) from the last hidden layer of the ResNet model used in PraNet.

    In the leftmost panel of Figure \ref{fig:polyp}, we report the long-term FNR for varying values of the localization parameter  $l$, targeting an FNR level $\alpha =0.1$. All methods converge rapidly to the desired FNR level, ensuring long-term risk control. The calibrated models are then tested on the hold-out data, and the FNR and average predicted set size are separately evaluated across different repositories. In the middle and right panels of Figure \ref{fig:polyp}, we report the average FNR and average prediction set size averaged over 10 trials.
    
    The model calibrated via ARC has a marginalized FNR error larger than the target value $\alpha $. Moreover, the FNR error is unevenly distributed across the different data repositories, ranging from $\text{FNR}=0.08$ for CVC-300 to $\text{FNR}=0.32$ for ETIS-LaribPolypDB. In contrast, L-ARC can equalize performance across repositories, while also achieving a test FNR closer to the target level. In particular, as illustrated in the rightmost panel, L-ARC improves the FNR for the most challenging subpopulation in the data by increasing the associated prediction set size, while maintaining a similar size for subpopulations that already have satisfactory performance.
   
    \subsection{Beam Selection}
    \label{sec:exp_beam_selection}
    \begin{figure}[htp]
    \begin{minipage}[b]{\textwidth} 
    \begin{subfigure}{0.25\textwidth}
    \includegraphics[width=\textwidth]{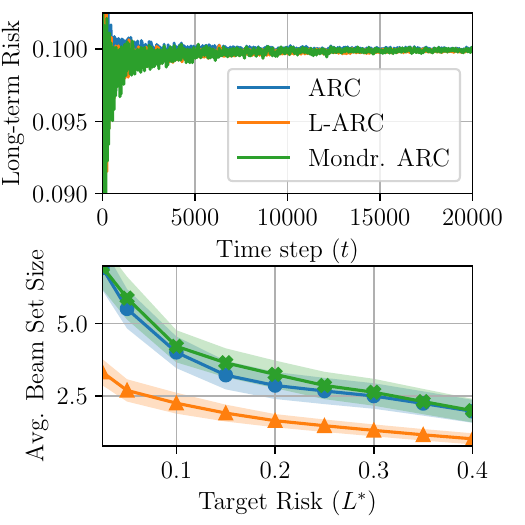}
    \end{subfigure}%
    \begin{subfigure}{0.75\textwidth}
    \includegraphics[width=\textwidth]{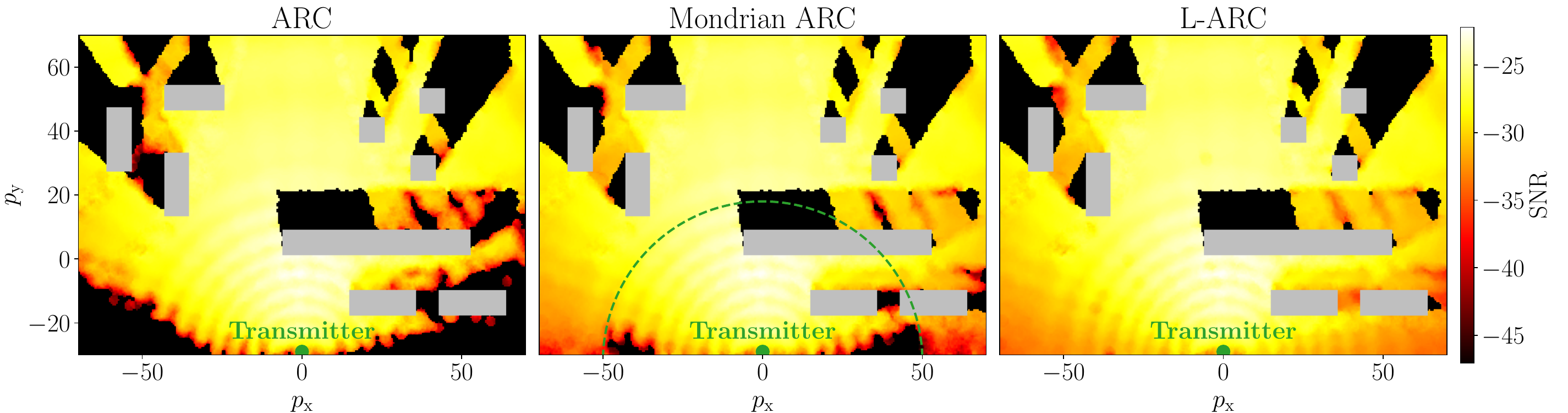}
    \vspace{0.1em}
    \end{subfigure}
    \caption{Long-term risk (left-top), average beam set size (left-bottom), and SNR level across the deployment area (right) for ARC, Mondrian ARC, and L-ARC. The transmitter is denoted as a green circle and obstacles to propagation are shown as grey rectangles. }
    \label{fig:beam-sweeping}
    \end{minipage}
    \end{figure}

	Motivated by the importance of reliable uncertainty quantification in engineering applications, we address the task of selecting location-specific beams for the initial access procedure in sixth-generation wireless networks \citep{ali2017millimeter}. Further details regarding the engineering aspects of the problem and the simulation scenario are provided in Appendix \ref{app:beamsweeping}. In the beam selection task, at each time $t$, the observed covariate corresponds to the location $X_t=[p_\mathrm{x},p_\mathrm{y}]$ of a receiver within the network deployment, where $p_\mathrm{x}$ and $p_\mathrm{y}$ represent the geographical coordinates. Based on the observed covariate, the transmitter chooses a set, denoted as $C_t\subseteq[1,\cdots,B_{\textrm{max}}]$, consisting of a subset of the $B_{\textrm{max}}$ available communication beams.
 
 Each communication beam $i$ is associated with a wireless link characterized by a signal-to-noise ratio $Y_{t,i}$, which follows an unknown distribution depending on the user's location $X_t$. We represent the vector of signal-to-noise ratios as $Y_t=[Y_{t,1},\dots,Y_{t,B_{\textrm{max}}}]\in\mathbb{R}^{B_{\textrm{max}}}$.
	For a set $C_t$, the transmitter sweeps over the beam set $C_t$, and the performance is measured by the ratio between the SNR obtained on the best beam in set $C_t$ and the best SNR on all the beams, i.e.,
	\begin{align}
		\mathcal{L}(C_t,Y_t)=L_t=1-\frac{\max_{i\in\mathcal{C}_t}Y_{t,i}}{\max_{i\in \{1,\dots,B_{\mathrm{max}}\}}Y_{t,i}}.
		\label{eq:SNR_regret}
	\end{align}
    Given an SNR predictor $\hat{Y}_t=f_{\mathrm{SNR}}(X_t)$ for all beams at location $X_t$, we consider sets that include only beams with a predicted SNR exceeding a threshold $g_t(X_t)$ as
    \begin{align}
       C(X_t,g_t)=\{i\in[1,\dots,B_{\textrm{max}}]:\hat{Y}_{t,i}>g_t(X_t)\}.
       \label{eq:beam-sweep-set}
    \end{align}
      In this setting, localization refers to the fair provision of service across the entire deployment area. As a benchmark, we thus also consider an additional calibration strategy that divides the deployment area into two regions: one encompassing all locations near the transmitter, which are more likely to experience high SNR levels, and the other including locations far from the transmitter. For each of these regions, we run two separate instances of ARC algorithms.  Inspired by the method introduced in \cite{bostrom2021mondrian} for offline settings, we refer to this baseline approach as Mondrian ARC.
    
   In the left panels of Figure \ref{fig:beam-sweeping}, we compare the performance of ARC, Mondrian ARC, and L-ARC with an RBF kernel with $l=10$, using a calibration data sequence of length $T=25000$. All methods achieve the target long-term SNR regret, but L-ARC achieves this result while selecting sets with smaller sizes, thus requiring less time for beam sweeping. Additionally, as illustrated on the right panel, thanks to the localization of the threshold function, L-ARC ensures a satisfactory communication SNR level across the entire deployment area. In contrast, both ARC and Mondrian ARC produce beam-sweep sets with uneven guarantees over the network deployment area.

    \section{Related Work}
    Our work contributes to the field of adaptive conformal prediction (CP), originally introduced by \citet{gibbs2021adaptive}. Adaptive CP extends traditional CP \citep{vovk2005algorithmic} to online settings, where data is non-exchangeable and may be affected by distribution shifts. This extension has found applications in reliable time-series forecasting \citep{xu2021conformal,zaffran2022adaptive}, control \citep{lekeufack2023conformal,angelopoulos2024conformal}, and optimization \citep{zhang2023bayesian,deshpande2024online}. Adaptive CP ensures that prediction sets generated by the algorithm contain the response variable with a user-defined coverage level on average across the entire time horizon. Recently, \citet{bhatnagar2023improved} proposed a variant of adaptive CP based on strongly adaptive online learning, providing coverage guarantees for any subsequence of the data stream. While their approach offers localized guarantees in time, L-ARC provides localized guarantees in the covariate space. More similar to our work is \citep{bastani2022practical}, which studies group-conditional coverage. Our work extends beyond coverage guarantees to a more general risk definition, akin to \cite{feldman2022achieving}. \citet{angelopoulos2024conformal} studied the asymptotic coverage properties of adaptive conformal predictions in the i.i.d. setting; and our work extends these results to encompass covariate shifts. Finally, the guarantee provided by L-ARC is similar to that of \citet{gibbs2023conformal}, albeit for an offline conformal prediction setting.

    \section{Conclusion and Limitations}
    \label{sec:conclusion}
    We have presented and analyzed L-ARC, a variant of adaptive risk control that produces prediction sets based on a threshold function mapping covariate information to localized threshold values. L-ARC can guarantee both worst-case deterministic long-term risk control and statistical localized risk control. Empirical analysis demonstrates L-ARC's ability to effectively control risk for different tasks while providing prediction sets that exhibit consistent performance across various data sub-populations. The effectiveness of L-ARC is contingent upon selecting an appropriate kernel function. Furthermore, L-ARC has memory requirements that grow with time due to the need to store the input data $\{X_t\}_{t>1}$ and coefficients (\ref{eq:coeffsLARC_1})-(\ref{eq:coeffsLARC}). These limitations of L-ARC motivate future work aimed at optimizing online the kernel function based on hold-out data \citep{kiyani2024conformal} or in an online manner \citep{angelopoulos2024conformal}, and at studying the statistical guarantees of memory-efficient variants of L-ARC \citep{kivinen2004online}.  

     \section{Acknowledgments}
    This work was supported by the European Union’s Horizon Europe project CENTRIC (101096379). The work of Osvaldo Simeone was also supported by the Open Fellowships of the EPSRC (EP/W024101/1) by the EPSRC project  (EP/X011852/1), and by Project REASON, a UK Government funded project under the Future Open Networks Research Challenge (FONRC) sponsored by the Department of Science Innovation and Technology (DSIT). We would also like to express our gratitude to Anastasios Angelopoulos for valuable insights on the technical content of the paper.
    
    \bibliography{biblio}
    \bibliographystyle{plainnat}
    \newpage
 	\appendix

    \section{Proof of Theorem \ref{th:iid}}
    \label{app:th2}
    We are interested in bounding the localized risk in (\ref{eq:formal_stat_guarantee}) of the threshold function (\ref{eq:time_averaged}) for all weighting functions in set $\mathcal{W}$ defined in (\ref{eq:covariate_shifts_set}).
    To study the limit in (\ref{eq:formal_stat_guarantee}) we first note that L-ARC update rule (\ref{eq:kernel_update}) corresponds to an online gradient descent step for a loss function $\ell(g,x,y)$, with respect to function $f(\cdot)$ and constant $c$ in function $g(\cdot)=f(\cdot)+c$ as in (\ref{eq:LARC_threshold}). {\color{black} In particular, interpreting the update rule (\ref{eq:constant_component})-(\ref{eq:kernel_update}) as a gradient descent step, we obtain that the partial derivatives of the loss function $\ell(g,x,y)$ evaluated at $g(\cdot)=f(\cdot)+c$  are }
    \begin{align}
        \nabla_f\ell(g)=\frac{\partial\ell(g,x,y)}{\partial f}(\cdot)&=(\alpha -\mathcal{L}(C(x,g),y))k(x,\cdot)+\lambda f(\cdot)\in\mathcal{H}\label{eq:gradient},\\
         \nabla_c\ell(g)=\frac{\partial\ell(g,x,y)}{\partial c}&=(\alpha -\mathcal{L}(C(x,g),y))\in\mathbb{R}\label{eq:gradient_2},
    \end{align}
    so that the first order approximation of the loss $\ell(g,x,y)$ around $g(\cdot)$  is given by
    \begin{align}
        \ell(g+\epsilon \delta_f,x,y)&\approx\ell(g,x,y)+\epsilon (\alpha -\mathcal{L}(C(x,g),y))\langle K_x,\delta_f\rangle+\epsilon\langle f,\delta_f \rangle\\
        \ell(g+\epsilon \delta_c,x,y)&\approx\ell(g,x,y)+\epsilon (\alpha -\mathcal{L}(C(x,g),y))\delta_c.
    \end{align}
    In order to study the convexity of the loss $\ell(g,x,y)$ in $g(\cdot)$,  we compute the the derivatives of (\ref{eq:gradient})-(\ref{eq:gradient_2}) with respect to $f(\cdot)$ and $c$. The derivative of (\ref{eq:gradient}) with respect to $f$ is the operator $A:\mathcal{H}\to\mathcal{H}$ satisfying
    \begin{align}
       A\delta_f&=\lim_{\epsilon\to 0} \frac{\nabla_f\ell(g+\epsilon\delta_f)-\nabla_f\ell(g)}{\epsilon} \nonumber\\
    &=-\lim_{\epsilon\to 0}\frac{\mathcal{L}(C(x,g),y)-\mathcal{L}(C(x,g+\epsilon\delta_f),y)}{\epsilon}K_x+\lambda\delta_f \nonumber\\
    &=-\Big\langle\frac{\partial\mathcal{L}(C(x,g),y)}{\partial g(x)}\frac{\partial g(x)}{\partial f},\delta_f\Big\rangle K_x+\lambda\delta_f \nonumber\\
    &=-\frac{\partial\mathcal{L}(C(x,g),y)}{\partial g(x)} \langle K_x,\delta_f\rangle K_x+\lambda\delta_f.
    \end{align}
    It follows that
    \begin{align}
         \langle f,A f\rangle=-\frac{\partial\mathcal{L}(C(x,g),y)}{\partial g(x)} f(x)^2+\lambda\norm{f}^2_{\mathcal{H}}.
    \end{align}
    Similarly, the derivative of (\ref{eq:gradient}) with respect to $c$ is the operator $B:\mathbb{R}\to\mathcal{H}$ is given by
    \begin{align}
        Bc&=-\frac{\partial\mathcal{L}(C(x,g),y)}{\partial g(x)}K_x c,
    \end{align}
    which satisfies 
    \begin{align}
        \langle f,Bc\rangle&=-\frac{\partial\mathcal{L}(C(x,g),y)}{\partial g(x)}f(x) c.
    \end{align}
    The derivative of (\ref{eq:gradient_2}) with respect to $c$ is given by 
    \begin{align}
        D c&=-\frac{\partial\mathcal{L}(C(x,g),y)}{\partial g(x)}c,
    \end{align}
    and the derivative with respect to to $f$ is the operator $C:\mathbb{\mathcal{H}}\to\mathcal{R}$ given by
    \begin{align}
        C \delta_f =-\frac{\partial\mathcal{L}(C(x,g),y)}{\partial g(x)}\langle K_x,\delta_f\rangle,
    \end{align}
    so that 
    \begin{align}
        \langle c, C f\rangle= -\frac{\partial\mathcal{L}(C(x,g),y)}{\partial g(x)}f(x) c.
    \end{align}
    {\color{black}
    From Assumption \ref{ass:unique_minimizer}, the inequality  $L'=-\mathbb{E}_{Y}\left[\frac{\partial\mathcal{L}(C(x,g),Y)}{\partial g(x)}|X=x\right]\geq \gamma>0$ holds. Thus, the second-order term of the approximation of $\mathbb{E}_{Y}\left[\ell(g,X,Y)|X=x\right]$ around $g(\cdot)\neq 0$ satisfies
    \begin{align}
       \mathbb{E}_{Y}\left[ \begin{bmatrix}
f & c
\end{bmatrix}\begin{bmatrix}
A & B \\
C & D 
\end{bmatrix}\begin{bmatrix}
f\\ 
c
\end{bmatrix}\Bigg| X=x\right]=&\mathbb{E}_{Y}\left[\langle f,A f\rangle+ \langle f,Bc\rangle+ \langle c, C f\rangle+\langle c, D c\rangle| X=x\right]\nonumber\\
=& L'f(x)^2+2L'f(x)c+L'c^2+\lambda\norm{f}^2_{\mathcal{H}}\nonumber\\
=& L'(f(x)+c)^2+\lambda\norm{f}^2_{\mathcal{H}}> 0.
    \end{align}
      We then conclude that the loss function $\mathbb{E}_{Y}\left[\ell(g,X,Y)|X=x\right]$ is strongly convex in $g(\cdot)$, and that the population loss minimizer
    \begin{align}
       g^{*}(\cdot)=f^{*}(\cdot)+c^*=\argmin_{g\in\mathcal{G}}\mathbb{E}_{X,Y}\left[\ell(g,X,Y)\right]
    \end{align}
    is unique. } For any covariate shift $w(\cdot)\in \mathcal{W}$ denote its components $f_w(\cdot)\in\mathcal{H}$ and $c_w\in \mathbb{R}$ such that $w(\cdot)=f_w(\cdot)+c_w$. From the first order optimality conditions, it holds that the directional derivatives with respect to $f_w(\cdot)$ and $c_w$ must satisfy
	\begin{align}
			\mathbb{E}_{X,Y}\left[\nabla_\epsilon\ell(g^{*}+\epsilon f_w,X,Y)|_{\epsilon=0}\right]=&\mathbb{E}_{X,Y}\left[(\alpha -\mathcal{L}(C_t(X,g^{*}),Y))f_w(X)+\lambda\langle f_w,f^{*}\rangle_{\mathcal{H}}\right]=0,\\
           \mathbb{E}_{X,Y}\left[\nabla_\epsilon\ell(g^{*}+\epsilon c_w,X,Y)|_{\epsilon=0}\right]=&\mathbb{E}_{X,Y}\left[(\alpha -\mathcal{L}(C_t(X,g^{*}),Y))c_w\right]=0,
	\end{align}
	which implies that for the optimal solution $g^{*}(\cdot)$ 
	\begin{align}
		\mathbb{E}_{X,Y}\left[\frac{w(X)}{\mathbb{E}_X[w(X)]}\mathcal{L}(C(X,g^{*}),Y)\right]= \alpha +\lambda\left\langle f^{*},\frac{f_w}{{\mathbb{E}_X[w(X)]}}\right\rangle_{\mathcal{H}}.
		\label{eq:marginal_cov}
	\end{align}
    Equality (\ref{eq:marginal_cov}) amounts to a localized risk control guarantee for the threshold $g^{*}(\cdot)$ for covariate shift in $w(\cdot)\in \mathcal{W}$. The following lemma states that the time-average L-ARC threshold function $\bar{g}_T(\cdot)$ defined in (\ref{eq:time_averaged}) converges to the population risk minimizer $g^*(\cdot)$.

    \begin{lemma}
    \label{lemma:conv}
        For any regularization parameter $\lambda>0$ and any learning rate sequence $\eta_t=\eta_1 t^{-1/2}<1/\lambda$, for some $\eta_1>0$, given a sequence $\{(X_t,Y_t)\}^T_{t=1}$ of i.i.d. samples from $P_{XY}$, the time-averaged threshold function (\ref{eq:time_averaged}) satisfies for any $\epsilon>0$
        \begin{align}
			\lim_{T\to\infty}\Pr[	\norm{g^{*}-\bar{g}_T}_{\infty}\geq \epsilon]=0 
            \label{eq:convergence_infinitynorm}
		\end{align}
    \end{lemma}

    \begin{proof}
      To prove convergence in probability, we need to show that the loss function $\ell(g,X,Y)$ is bounded. To this end, we first show that $\ell(g,X,Y)$ is Lipschitz in $g(\cdot)$ by studying the norm of the derivatives (\ref{eq:gradient})-(\ref{eq:gradient_2}). For $g_t(\cdot)=f_t(\cdot)+c_t$ returned by the update rule (\ref{eq:kernel_update}), the gradient with respect to $f_t(\cdot)$ satisfies
       \begin{align}
            \norm{\frac{\partial\ell(g_t,x,y)}{\partial f}(\cdot)}_{\mathcal{H}}&=\norm{(\alpha -\mathcal{L}(C(x,g_t),y)k(x,\cdot)+\lambda f_t(\cdot)}_{\mathcal{H}}\nonumber\\
            &\leq B\sqrt{\kappa}+\lambda \norm{f_t(\cdot)}_{\mathcal{H}}\leq  2B\sqrt{\kappa},
        \end{align}
        where the first inequality follows from the boundedness on the kernel (Assumption \ref{ass:lower_bound_k}) and the boundedness on the loss (Assumption \ref{ass:loss_bounded}), while the last follows from Proposition \ref{prop:bound_rkhs}. The gradient with respect to $c$ can be similarly bounded as
        \begin{align}
            \left|\frac{\partial\ell(g_t,x,y)}{\partial c}(\cdot)\right|&=\left|(\alpha -\mathcal{L}(C(x,g_t),y)\right|\leq B.
        \end{align}
        From the mean value theorem it follows that for $g(\cdot)=f(\cdot)+c$ and $g'(\cdot)=f'(\cdot)+c'$
        \begin{align}
           | \ell(g,X,Y)-\ell(g',X,Y)|\leq \left(2B\sqrt{k}+B\right)\norm{f-f'}_{\mathcal{H}}+B|(c-c')|.
        \end{align}
        Since L-ARC returns functions $f_t(\cdot)$ with bounded RKHS norm and infinity norm (Proposition \ref{prop:bound_rkhs}), and thresholds function $g_t(\cdot)$ with bounded in infinity norm (Proposition \ref{prop:bound_f}), we conclude that there exists a finite $\ell_{max}<\infty$ such that $|\ell(g,X,Y)|\leq \ell_{max}$.  Given that the loss is bounded we can apply \cite[Theorem 4]{kivinen2004online} and obtain that for the threshold $\{g_t(\cdot)\}_{t\geq 1}$ returned by L-ARC and the population loss minimizer $g^*(\cdot)$ it holds
        \begin{align}
          \frac{1}{T}\sum^{T}_{t=1}		\ell(g_t,X_t,Y_t)  \leq \frac{1}{T}\sum^{T}_{t=1}\ell(g^*,X_t,Y_t) + B^2\kappa^2(2\kappa^2+1)^2\left(\frac{2}{\sqrt{T}}\left(2\eta_0+\frac{1}{\eta_0 \lambda^2}\right)+\frac{1}{2\eta_0\lambda^2 T}\right)\hspace{-0.3em}.
          \label{eq:regret_bound}
        \end{align}
        By Hoeffding's inequality the empirical average on the right-hand side of (\ref{eq:regret_bound}) converges to its expected value. Formally, we have that with probability at least $1-\delta$ with respect to the sequence $\{(X_t,Y_t)\}^T_{t=1}$ 
        \begin{align}
            \left|\frac{1}{T}\sum^{T}_{t=1}\ell(g^*,X_t,Y_t)- \mathbb{E}_{X,Y}[\ell(g^*,X,Y)]\right|\leq \ell_{max}\sqrt{\frac{2}{T}\log\left(\frac{1}{\delta}\right)}.
        \end{align}
        Similarly, by \cite[Theorem 2]{cesa2004generalization} the empirical risk on the left-hand side of (\ref{eq:regret_bound}) converges to the population risk of the time-averaged solution (\ref{eq:time_averaged}). With probability at least $1-\delta$ with respect to the sequence of samples $\{(X_t,Y_t)\}^T_{t=1}$, it holds
        \begin{align}
			\mathbb{E}_{X,Y}\left[\ell(\bar{g}_T,X,Y)\right]\leq  \frac{1}{T}\sum^{T}_{t=1}		\ell(g_t,X_t,Y_t) + \ell_{max}\sqrt{\frac{2}{T}\log\left(\frac{1}{\delta}\right)}
		\end{align}
        Combining the two inequalities, with probability at least $1-2\delta$ with respect to $\{(X_t,Y_t)\}^T_{t=1}$,
		\begin{align}
			\mathbb{E}_{X,Y}\left[\ell(\bar{g}_T,X,Y)-\ell(g^{*},X,Y)\right]\leq & B^2\kappa^2(2\kappa^2+1)^2\left(\frac{2}{\sqrt{T}}\left(2\eta_0+\frac{1}{\eta_0 \lambda^2}\right)+\frac{1}{2\eta_0\lambda^2 T}\right)\nonumber\\
   &+2\ell_{max}\sqrt{\frac{2}{T}\log\left(\frac{1}{\delta}\right)}
			\label{eq:concentration}
		\end{align}
		Since the $\mathbb{E}_{X,Y}[\ell(g,X,Y)]$ is strongly convex there exists a value $\gamma>0$ such that the second order approximation of $\mathbb{E}_{X,Y}[\ell(g,X,Y)]$ at $g^{*}(\cdot)$ satisfies 
		\begin{align}
			\frac{\gamma}{2}\left(\norm{f^{*}-\bar{f}_T}_{\mathcal{H}}+(c^*-\bar{c}_T)\right)^2 \leq  \mathbb{E}_{X,Y}[\ell(\bar{g}_T,X,Y)-\ell(g^{*},X,Y)]
			\label{eq:taylor_approx}
		\end{align}
		Combining (\ref{eq:taylor_approx}) and (\ref{eq:concentration}), and leveraging $\norm{f}_\infty\leq \sqrt{\kappa}\norm{f}_{\mathcal{H}}$, which follows from the Assumption \ref{ass:lower_bound_k}, we conclude that with probability $1-2\delta$
		\begin{align}
			\norm{g^{*}-\bar{g}_T}_{\infty}\leq&\sqrt{ \frac{2 B^2\kappa^2(2\kappa^2+1)^2}{\gamma(\kappa+1)}\hspace{-0.2em}\left(\frac{2}{\sqrt{T}}\hspace{-0.2em}\left(2\eta_0+\frac{1}{\eta_0 \lambda^2}\right)\hspace{-0.2em}+\hspace{-0.2em}\frac{1}{2\eta_0\lambda^2 T}\right)\hspace{-0.2em}+\hspace{-0.2em}\frac{4\ell_{max}}{\gamma(\kappa+1)}\sqrt{\frac{2}{T}\log\left(\frac{1}{\delta}\right)}}.
		\end{align}
		Choosing $\delta=\frac{1}{T}$, for any $\epsilon>0$, it holds
		\begin{align}
			\lim_{T\to\infty}\Pr[	\norm{g^{*}-\bar{g}_T}_{\infty}\geq \epsilon]=0 .
		\end{align}
    \end{proof}
    By itself, the convergence of the threshold function $\bar{g}_T(\cdot)$ to the population risk minimizer $g^{*}(\cdot)$ is not sufficient to provide localized risk control guarantees for L-ARC time-averaged solution. However,  under the additional loss regularity assumption in Assumption \ref{ass:left_continuity}, we can show that set predictor $C(X,\bar{g}_T)$ enjoys conditional risk control for $T\to\infty$.
    
    Having assumed that the loss $\mathcal{L}(C(x,g),y)$ is left-continuous and decreasing for larger prediction sets (Assumption \ref{ass:loss_bounded} and \ref{ass:left_continuity}), for any $\delta'> 0$ there exists $\epsilon>0$ such that for $g(\cdot)$ such that $\norm{g^{*}-g}_\infty\leq \epsilon$ it holds
        \begin{align}
            \mathcal{L}\left(C\left(X,g\right),Y\right)\leq \mathcal{L}\left(C\left(X,g^{*}\right),Y\right)+\delta'.
        \end{align}
        For such $g(\cdot)$ the following inequality holds
        \begin{align}
            \max_{w\in\mathcal{W}}	\mathbb{E}\left[\frac{w(X)}{\mathbb{E}[w(X)]}\mathcal{L}\left(C\left(X,g\right),Y\right)\right]\leq \max_{w\in\mathcal{W}}	\mathbb{E}\left[\frac{w(X)}{\mathbb{E}[w(X)]}\mathcal{L}\left(C\left(X,g^{*}\right),Y\right)\right]+\delta'.
            \label{eq:intermedia_1}
        \end{align}
         As stated in Lemma \ref{lemma:conv}, we can always find $T$ large enough, such that $\norm{g^{*}-\bar{g}_T}_\infty\leq \epsilon$ with arbitrary large probability. This implies, that for any $\delta'>0$ and $w\in\mathcal{W}$,
         \begin{align}
            \lim_{T\to\infty}	\mathbb{E}\left[\frac{w(X)}{\mathbb{E}[w(X)]}\mathcal{L}\left(C\left(X,\bar{g}_T\right),Y\right)\right]&\leq 	\mathbb{E}\left[\frac{w(X)}{\mathbb{E}[w(X)]}\mathcal{L}\left(C\left(X,g^{*}\right),Y\right)\right]+\delta'\label{eq:th_final_step_1}\\
            &\leq \alpha +\lambda\left\langle f^{*},\frac{f_w}{{\mathbb{E}_X[w(X)]}}\right\rangle_{\mathcal{H}}+\delta'\label{eq:th_final_step_2}\\
            &\leq  \alpha + \kappa B\frac{\norm{f_w}_{\mathcal{H}}}{{\mathbb{E}_X[w(X)]}}+\delta',
            \label{eq:app:final_th1}
        \end{align}
		where the inequality (\ref{eq:th_final_step_1}) follows from (\ref{eq:intermedia_1}), the inequality (\ref{eq:th_final_step_2}) follows from  (\ref{eq:marginal_cov}) and  the inequality (\ref{eq:app:final_th1}) from Proposition \ref{prop:bound_rkhs}.
        \section{Proof of Theorem \ref{th1}}
	\label{app:th1}
	We are interested in bounding the absolute difference between the cumulative loss value incurred by the set predictors $\{C(g_t,X_t)\}^T_{t=1}$ produced by L-ARC $(\ref{eq:kernel_update})$ and the target reliability level $\alpha $, i.e.,
	\begin{align}
		\left|\frac{1}{T}\sum^T_{t=1}(\underbrace{\mathcal{L}(C_t,Y_t)}_{L_t}-\alpha )\right|.	
	\end{align}
	From Assumption \ref{ass:lower_bound_k} and having assumed $\norm{X_t}\leq D$ for $t\geq 1$, it follows that 
	\begin{align}
		\lim_{\norm{x}\to\infty}k(X_t,x)= 0.
	\end{align}
	A bound on the cumulative risk can then be obtained by bounding
	\begin{align}
		\norm{\frac{1}{T}\sum^T_{t=1}(L_t-\alpha )(k(X_t,\cdot)+1)}_\infty,
		\label{eq:target_func_to_bound}
	\end{align}
    where for a function $f:\mathcal{X}\to \mathbb{R}$, the infinity norm $\norm{f}_{\infty}$ is defined as $\max_{x\in\mathcal{X}}|f(x)|$.
    In fact, from (\ref{eq:target_func_to_bound}) we directly obtain a bound on the cumulative risk 
	\begin{align}
		\left|\frac{1}{T}\sum^T_{t=1}(L_t-\alpha )\right|&=\lim_{\norm{x}\to\infty}\left|\frac{1}{T}\sum^T_{t=1}(L_t-\alpha )(k(X_t,x)+1)\right|\leq \norm{\frac{1}{T}\sum^T_{t=1}(L_t-\alpha )(k(X_t,\cdot)+1)}_\infty\hspace{-0.5em}.
		\label{eq:mistake_limit}
	\end{align}
	To this end, we first note that functions $\{f_t(\cdot)\}_{t\in\mathbb{N}}$ generated by (\ref{eq:kernel_update}) have bounded RKHS norm and are smooth.
	\begin{proposition}
		\label{prop:bound_rkhs}
		For every $t\geq 1$, we have the inequalities $\norm{f_t}_{\mathcal{H}}\leq \frac{B \sqrt{\kappa}}{\lambda}$ and $\norm{f_t}_{\infty}\leq \frac{\kappa B}{\lambda}$.
	\end{proposition}
	\begin{proof}
        {\color{black}
        The proof is by induction, with the base case $\norm{f_1(\cdot)}_\mathcal{H}\leq B\sqrt{\kappa}/\lambda$
		being satisfied as $f_1(\cdot)=0$. The induction step is given as
		\begin{align}
			\norm{f_{t+1}}_{\mathcal{H}}&=\norm{(1-\lambda\eta_t)f_t-\eta_t(\alpha -L_t)k(x_t,\cdot)}_{\mathcal{H}} \label{eq:prop_1_1}\\
			&\leq \norm{(1-\lambda\eta_t)f_t}_{\mathcal{H}}+ \norm{\eta_t(\alpha -L_t)k(x_t,\cdot)}_{\mathcal{H}}\label{eq:prop_1_2}\\
			&\leq (1-\lambda\eta_t)\norm{f_t}_{\mathcal{H}}+\eta_tB\sqrt{\kappa}\label{eq:prop_1_3}\\
            &\leq \frac{B\sqrt{\kappa}}{\lambda},\label{eq:prop_1_4}
		\end{align}
		where the equality (\ref{eq:prop_1_1}) follows from the update rule (\ref{eq:kernel_update}); the inequality (\ref{eq:prop_1_2}) from the properties of the norm, the inequality (\ref{eq:prop_1_3}) from Assumption \ref{ass:lower_bound_k} and \ref{ass:loss_bounded}, and the inequality (\ref{eq:prop_1_4}) from the induction hypothesis $\norm{f_t}_{\mathcal{H}}\leq \frac{B\sqrt{\kappa}}{\lambda}$.}
    \end{proof}

  \begin{proposition}
		\label{prop:lipshitz}
		For $t\geq 1$ and any $(x,x')\in \mathcal{X}\times \mathcal{X}$  we have
        \begin{align}
            |f(x)-f(x')|\leq \frac{B \sqrt{2\rho \kappa D}}{\lambda}.
        \end{align}
	\end{proposition}
    \begin{proof}
    Denote the evaluation function at $x$ as $K_x=k(x,\cdot)$. From Proposition \ref{prop:bound_rkhs} and the Lipschitz continuity assumed in Assumption \ref{ass:lower_bound_k}, it follows that 
    \begin{align}
        |f(x)-f(y)|&=|\langle f,K_x\rangle_{\mathcal{H}}-\langle f,K_y\rangle_{\mathcal{H}}|\nonumber\\
        &=|\langle f,K_x-K_y\rangle_{\mathcal{H}}|\nonumber\\
        &\leq \norm{f}_\mathcal{H} \norm{K_x-K_y}_\mathcal{H}\nonumber\\
        &= \norm{f}_\mathcal{H} \sqrt{k(x,x)+k(y,y)-2k(x,y)}\nonumber\\
        &\leq \norm{f}_\mathcal{H} \sqrt{2\rho \norm{x-y}}\nonumber\\
        &\leq \frac{2B \sqrt{\rho \kappa D}}{\lambda} \label{eq:lipshitz}
    \end{align}
    where the first inequality follows from Cauchy–Schwarz inequality, the second from the Lipschitz continuity of the kernel, and the last one from Proposition \ref{prop:bound_rkhs} together with $\norm{x}\leq D$ for $x\in\mathcal{X}$.
    \end{proof}
    Leveraging the above characterization of the function $f_t(\cdot)$ returned by L-ARC, we now show that the threshold function $g_t(\cdot)$ has maximum and minimum values that are uniformly bounded.
	\begin{proposition}
		\label{prop:bound_f}
		For every $t\geq 1$ and $x\in\mathcal{X}$ we have $g_t(x)\in [G_{\textrm{min}},G_{\textrm{max}}]$ with
		\begin{align}
			G_{\textrm{max}}= S_{\textrm{max}}+\frac{2B \sqrt{\rho \kappa D}}{\lambda}+\eta_{0}B (2\kappa+1)
            \label{eq:ut}
		\end{align}
		and
		\begin{align} 
			G_{\textrm{min}}= -\frac{2B \sqrt{\rho \kappa D}}{\lambda}-\eta_{0}B (2\kappa+1). 
            \label{eq:lt}
		\end{align}
	\end{proposition}
	\begin{proof}
		We now prove the upper bound (\ref{eq:ut}). The proof is by contradiction and it start by assuming that there exists a $t>1$ and $x\in\mathcal{X}$ such that $g_t(x)\geq G_{\textrm{max}}$ while $ g_{t'}(\cdot)< G_{\textrm{max}}$ for all $t'<t$. From the update rule (\ref{eq:kernel_update}) we have that
		\begin{align}
            g_{t-1}(x)	&=g_{t}(x) +\eta_{t-1}(\alpha -L_t)(k(X_{t-1},x)+1)-\lambda\eta_{t-1}f_{t-1}(x)\nonumber\\
            &\geq G_{\textrm{max}}-\eta_{0}B (\kappa+1)-\lambda\eta_0|f_{t-1}(x)|\nonumber\\
            &\geq  G_{\textrm{max}}-\eta_{0}B (2\kappa+1).
		\end{align}
		From Proposition \ref{prop:lipshitz} we also have
		\begin{align}
			g_{t-1}(X_{t-1})&\geq g_{t-1}(x)-\frac{2B \sqrt{\rho \kappa D}}{\lambda}\geq   G_{\textrm{max}}-\eta_{0}B (2\kappa+1)-\frac{2B \sqrt{\rho \kappa D}}{\lambda}\geq S_{\textrm{max}},
		\end{align}
		where the last inequality follows from $G_{\textrm{max}}$ being defined as (\ref{eq:ut}).
  From Assumption \ref{ass:nc_score}, for all $x\in \mathcal{X}$,
		\begin{align}
			g_{t-1}(X_{t-1})\geq S_{\textrm{max}} \implies \alpha \geq L_{t-1} \implies g_t(x)\leq (1-\lambda\eta_{t-1})g_{t-1}(x)\leq G_{\textrm{max}},
		\end{align}
		which contradicts with the original assumption that there exists $x$ such that $g_t(x)\geq G_{\textrm{max}}$.

         The proof of the lower bound (\ref{eq:lt}) follows similarly. Assume there exists $t>1$ and $x\in\mathcal{X}$ such that $	g_t(x)\leq G_{\textrm{min}}$ while $g_{t'}(\cdot)>G_{\textrm{min}}$ for $t'<t$.  From the update rule (\ref{eq:kernel_update}) we have that
		\begin{align}
			g_{t-1}(x)	&=g_{t}(x) +\eta_{t-1}(\alpha -L_t)(k(X_{t-1},x)+1)-\lambda\eta_{t-1}f_{t-1}(x)\nonumber\\
            &\leq G_{\textrm{min}} +\eta_{0}B (\kappa+1)+\lambda\eta_{0}|f_{t-1}(x)|\nonumber\\
            &\leq  G_{\textrm{min}}+\eta_{0}B (2\kappa+1)
		\end{align}
		From Proposition \ref{prop:lipshitz} we also have
		\begin{align}
			g_{t-1}(X_{t-1})&\leq g_{t-1}(x)+\frac{2B \sqrt{\rho \kappa D}}{\lambda}\leq  G_{\textrm{min}}+\eta_{0}B (2\kappa+1)+\frac{2B \sqrt{\rho \kappa D}}{\lambda}\leq 0
		\end{align}
	   where the last inequality follows from $G_{\textrm{min}}$ being defined as (\ref{eq:lt}). From Assumption \ref{ass:nc_score}, for all $x\in \mathcal{X}$,
		\begin{align}
			g_{t-1}(X_{t-1})\leq 0 \implies L_{t-1}\geq \alpha  \implies g_t(x)\geq(1-\lambda\eta_{t-1})g_{t-1}(x)\geq G_{\textrm{min}}
		\end{align}
		which contradicts the assumption that there exists $\min_{x\in\mathcal{X}}	g_t(x)\leq G_{\textrm{min}}$.
	\end{proof}
	Having established an upper and lower bound on the maximum value of the function $g_t(\cdot)$ generated by (\ref{eq:kernel_update}) we can now bound (\ref{eq:target_func_to_bound}). Define $\Delta_t=\eta^{-1}_t-\eta^{-1}_{t-1}$ and $\Delta_1=\eta^{-1}_1$ and note that
	\begin{align}
		\left|\frac{1}{T}\sum^T_{t=1}(L_t-\alpha )\right|\leq& \max_{x\in\mathcal{X}}\left|{\frac{1}{T}\sum^T_{t=1}(L_t-\alpha )(k(X_t,x)+1)}\right|\nonumber\\
  =&\norm{\frac{1}{T}\sum^T_{t=1}\left(\sum^t_{r=1}\Delta_r\right)\eta_t(L_t-\alpha )(k(X_t,\cdot)+1)}_{\infty}\nonumber\\
		=&\norm{\frac{1}{T}\sum^T_{r=1}\Delta_r \left(\sum^T_{t=r}\eta_t(L_t-\alpha )(k(X_t,\cdot)+1)\right)}_{\infty}\nonumber\\
		=&\norm{\frac{1}{T}\sum^T_{r=1}\Delta_r \left(\sum^T_{t=r}f_{t+1}+c_{t+1}-(1-\lambda\eta_t)f_t-c_t\right)}_{\infty}\nonumber\\
		=&\norm{\frac{1}{T}\sum^T_{r=1}\Delta_r \left(g_{T+1}-g_r+\lambda\sum^T_{t=r}\eta_tf_t\right)}_{\infty}\nonumber\\
		\leq& \norm{\frac{1}{T}\sum^T_{r=1}\Delta_r \left(g_{T+1}-g_r\right)}_\infty+\norm{\frac{\lambda}{T}\sum^T_{r=1}\Delta_r\sum^T_{t=r}\eta_tf_t}_{\infty}\nonumber\\
		\leq&  \underbrace{\frac{1}{T}\sum^T_{r=1}\Delta_r \norm{g_{T+1}-g_r}_{\infty}}_{:={\color{black}E_1}}+\underbrace{\frac{\lambda}{T}\sum^T_{t=1} \norm{f_t}_{\infty}}_{:={\color{black}E_2}}.
	\end{align}
	The first term can be bounded based on Proposition (\ref{prop:bound_f}) as
	\begin{align}
		{\color{black}E_1}&\leq \frac{1}{T}\max_{r}\norm{g_{T+1}-g_r}_{\infty}\sum^T_{r=1}\Delta_r= \frac{1}{\eta_T T}\left(S_{\textrm{max}}+\frac{4B \sqrt{\rho \kappa D}}{\lambda}+2\eta_{0}B (2\kappa+1) \right),
	\end{align}
	and similarly, for the second term, we have
	\begin{align}
		{\color{black}E_2}&\leq \frac{\lambda}{T}\sum^T_{t=1}\frac{\kappa B}{\lambda}=\kappa B.
	\end{align}
	Fix a decreasing learning rate $\eta_t=\eta_0 t^{-\omega}$ and a regularization parameter $\lambda=\lambda_0T^{-\xi}$, then the ${\color{black}E_1}$ becomes
	\begin{align}
		{\color{black}E_1}= \frac{S_{\textrm{max}}}{\eta_0 T^{1-\omega}}+\frac{4B \sqrt{\rho \kappa D}}{\eta_0\lambda T^{1-\omega}}+\frac{2B (2\kappa+1)}{ T^{1-\omega}}
	\end{align}
	For any $\omega<1$, it follows 
	\begin{align}
	  \lim_{T\to \infty}	\left|\frac{1}{T}\sum^T_{t=1}(L_t-\alpha )\right|= \kappa B.
	\end{align}

\newpage	
\section{Additional Experiments}
\subsection{Beam Selection}
\subsubsection{Simulation Details}
\label{app:beamsweeping}
    \begin{figure}[ht]
		\centering
		\includegraphics[width=0.4\textwidth]{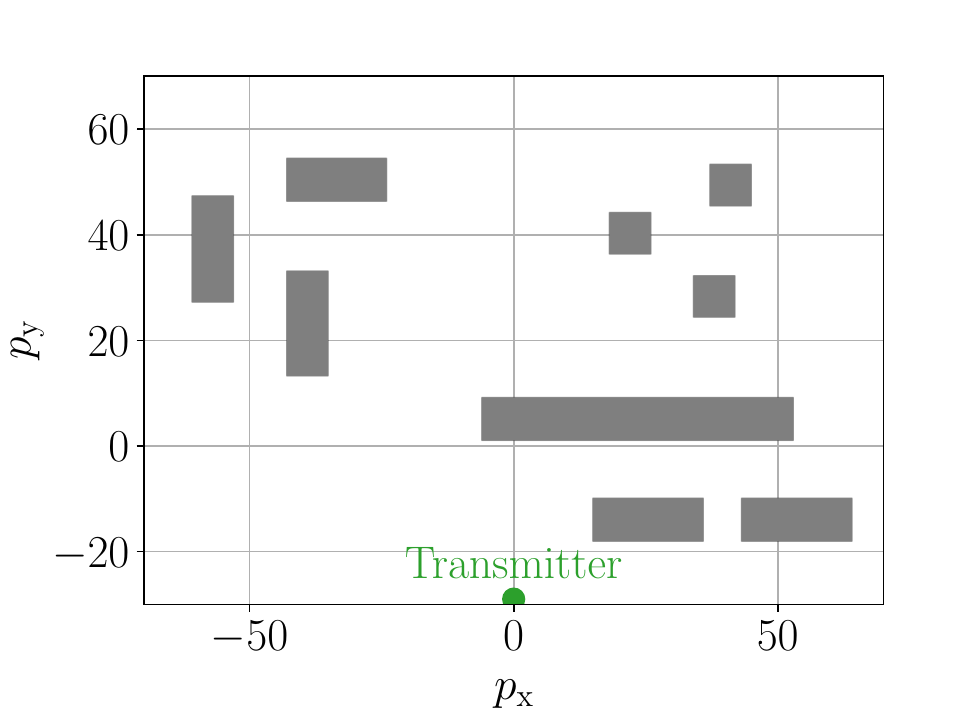}
		\caption{Network deployment assumed in the simulations. A single transmitter (green circle) communicates with receivers that are uniformly distributed in a scene containing multiple buildings (grey rectangles).}
		\label{fig:deployment}
	\end{figure}
 
	For the beam selection experiment, we consider the network deployment depicted in Figure \ref{fig:deployment}, in which a transmitter (green circle) communicates with users in an urban environment with multiple buildings (grey rectangles). We assume that communication occurs at a frequency $f_c=2.14$ GHz and that the transmitter is equipped with $N_t=8$ transmitting antennas while receiving users have single-antenna equipment. 
    The transmitter adopts a discrete Fourier transform beamforming codebook of size $B_{\mathrm{max}}=11$, with each beam $b_i$ given by
	\begin{align}
		b_i= \frac{1}{\sqrt {N_t}}[1,e^{j2\pi\frac{2\pi i }{B_{\mathrm{max}}}},\dots,e^{j(N_t-1)\frac{2\pi i }{B_{\mathrm{max}}}}]\in \mathbb{C}^{N_t}, \quad \text{for }i \in \{0,\dots,B_{\mathrm{max}}-1\},
	\end{align} 
    where $j=\sqrt{-1}$.
 The wireless channel response $h^R\in\mathbb{C}^{N_t}$ between the transmitter and a receiver located at $X_t=[p_\mathrm{x},p_\mathrm{y}]\in\mathbb{R}^2$,  is modeled using Sionna ray-tracer \citep{hoydis2023sionna}, and we account for small scale fading using a Rayleigh noise model \citep{goldsmith2005wireless}. The resulting channel vector is distributed as
	\begin{align}
		h_t\sim h^{R}(X_t)+Rayleigh(\sigma).
	\end{align}
    where $h^{R}(X_t)$ is the ray tracer output and $Rayleigh(\sigma)$ is a Rayleigh distributed random variable with parameter $\sigma=10^{-4}$.
    Assuming unit power transmit symbols and receiver noise, for a channel vector $h_t$ the communication signal-to-noise ratio (SNR) obtained using the beamformer $b_i$ is given by
    \begin{align}
        Y_{t,i}=h_t^{\mathrm{T}}b_i.
    \end{align} 
    Beam sets are obtained calibrating an SNR predictor $\hat{Y}_t=f_{\mathrm{SNR}}(X_t)$ realized using a 3-layer fully connected neural network that is trained on $2500$ samples with the user location generated uniformly at random within the deployment area.
    \subsubsection{Effect of the Length Scale}
     \begin{figure}[ht]
		\centering
		\includegraphics[width=\textwidth]{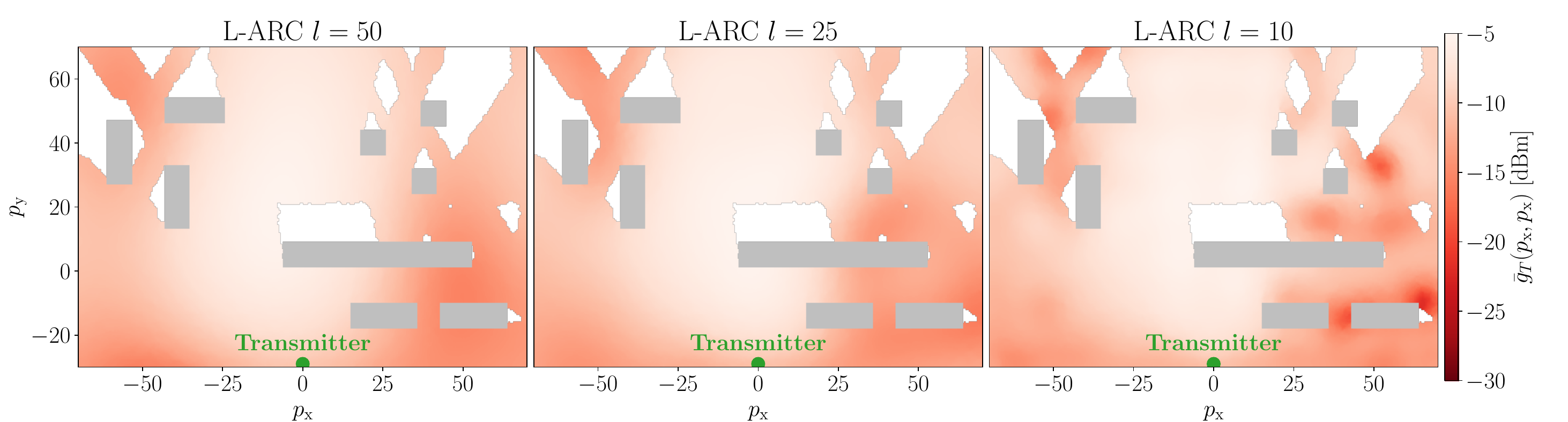}
		\caption{Time-averaged threshold function $\bar{g}_T$ for different values of localization parameter $l$.}
		\label{fig:thresholds}
	\end{figure}

   In Figure \ref{fig:thresholds}, we study the effect of the length scale $l$ of the kernel function on the time-averaged threshold function $\bar{g}_T(X)$ returned by L-ARC. We report the value of the L-ARC time-averaged threshold, $\bar{g}_T(X)$, in (\ref{eq:time_averaged}), for the same experimental set-up as in Section \ref{sec:exp_beam_selection}, and for increasing localization of the kernel function. As the length scale parameter $l$ decreases, corresponding to a more localized kernel, the value of the threshold is allowed to vary more across the deployment area. In particular, the threshold function reduces its value around areas where the beam selection problem becomes more challenging, such as building clusters, in order to create larger beam selection sets.
\subsection{Image Classification with Calibration Requirements}
\label{app:exp_condidence}
\begin{figure}[htp]
    \begin{minipage}[b]{\textwidth} 
    \begin{subfigure}{0.32\textwidth}
    \includegraphics[width=\textwidth]{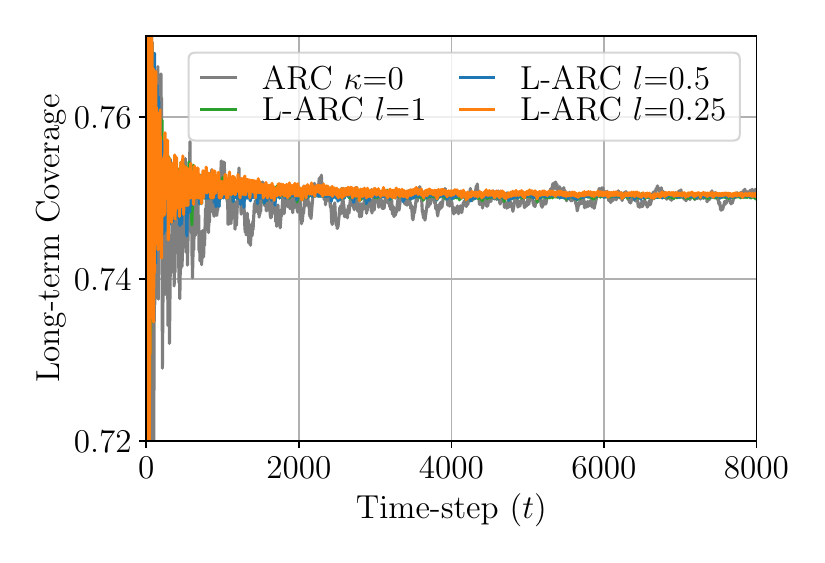}
    \end{subfigure}%
    \begin{subfigure}{0.64\textwidth}
    \includegraphics[width=\textwidth]{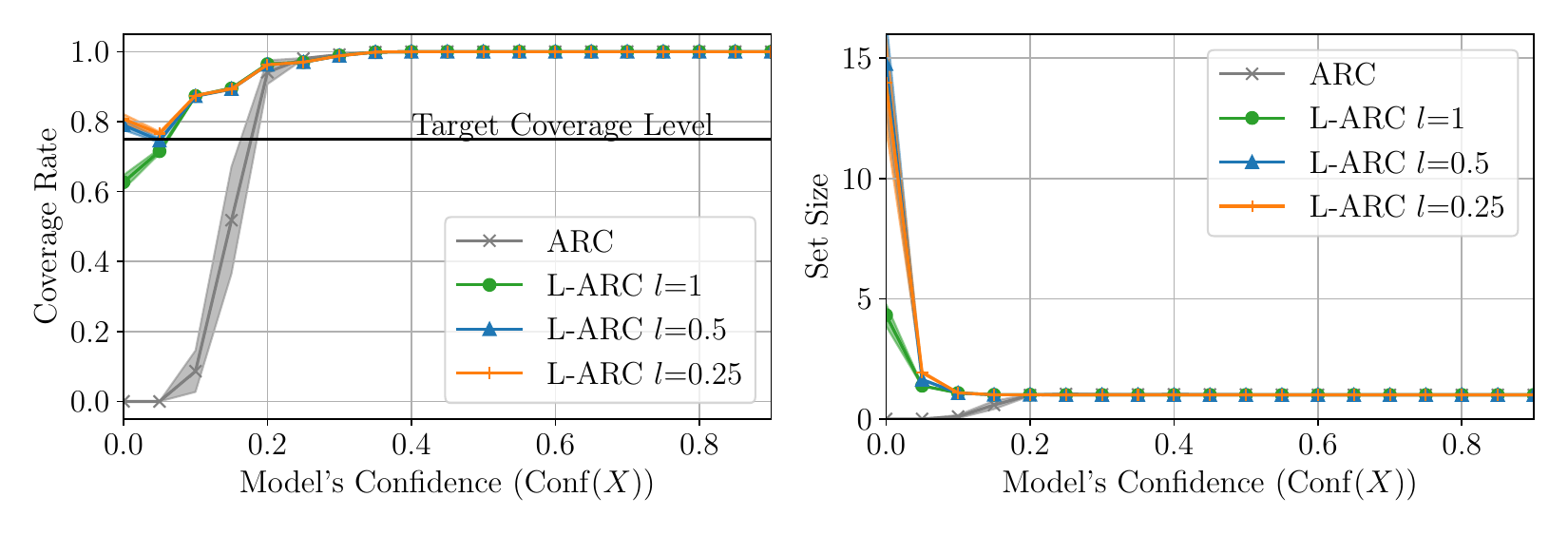}
    \end{subfigure}
    \caption{Long-term coverage (left), coverage rate (center), and prediction set size (right) versus model's confidence for ARC and L-ARC for different values of the localization parameter $l$. }
    \label{fig:image-classification-uncertainty}
    \end{minipage}
    \end{figure}
    In this section, we consider an image classification task under calibration requirements based on the fruit-360 dataset \citep{muresan2018fruit}. For this problem, the feature vector $X_t$ is an image of size $100\times100$, and the corresponding label $Y_t\in\mathcal{Y}=\{1,\dots,130\}$ is one of 130 types of fruit, vegetable, or nut in image $X_t$. We study the online calibration of a pre-trained ResNet18 model \citep{he2016deep}. For an input image $X_t$, the prediction set is obtained from the model's predictive distribution $\hat{p}(y|X_t)$ as 
     \begin{align}
       C(X_t,g_t)=\{y\in\mathcal{Y}:\hat{p}(y|X_t)>g_t(X_t)\},
       \label{eq:image-set}
    \end{align}
    and we target the miscoverage loss (\ref{eq:cov_loss}) with a target miscoverage rate $\alpha =0.25$.
    In order to capture calibration requirements, we impose coverage constraints that are localized in the model's confidence. The model's confidence indicator is given by the maximum value of the model's predictive distribution $\hat{p}(y|X_t)$, i.e.,
    \begin{align}
        \mathrm{Conf}(X_t)=\max_{y\in\mathcal{Y}}\hat{p}(y|X_t).
        \label{eq:model-conf}
    \end{align}
	Accordingly, we run ARC and L-ARC calibration with a sequence of $T=8000$ samples and we instantiate L-ARC using the exponential kernel $k(x,x')=\kappa \exp(-\norm{\phi(x)-\phi(x')}^2/l)$, where the feature vector is given by the model's uncertainty, i.e., $\phi(x)=\mathrm{Conf}(x)$.
 
 In the left-most panel of Figure \ref{fig:image-classification-uncertainty} we report the long-term coverage of ARC and L-ARC for an increasing level of localization obtained by decreasing the length scale $l$. All methods guarantee long-term coverage. In the middle panel, we use hold-out data to evaluate the coverage of the calibrated model conditioned on the model's confidence level. For small length scale $l$, L-ARC yields prediction sets that satisfy the coverage requirement across different levels of the model's confidence. In contrast, ARC, due to its inability to adapt the threshold function, has a large miscoverage rate for small model confidence levels. As illustrated in the right panel, this is achieved by producing a larger set size when the model's confidence is low.

\subsection{On the memory efficiency of L-ARC}
\label{sec:mem_eff}
In a manner similar to  \citep{kivinen2004online}, it is possible to obtain a memory-efficient version of L-ARC that adopts a truncated version of L-ARC threshold (\ref{eq:threshold}) given by
\begin{align}
	g_{t+1}(\cdot)=\hspace{-2em}\sum^{t}_{i=\max\{1,t-M_{\text{max}}\}}\hspace{-2em}a^i_{t+1} k(X_i,\cdot)+c_{t+1}.
	\label{eq:threshold_truncated}
\end{align}
Unlike the threshold (\ref{eq:threshold}), which has a \textit{linear} memory requirement, the truncated version (\ref{eq:threshold_truncated}) requires a \emph{constant} memory and computational load that are proportional to the number of coefficients $M_{\text{max}}$.
It is known that in online non-parametric learning, there exists a trade-off between memory efficiency and performance. In the following, we empirically study the trade-off between the localized risk control of L-ARC and its memory requirements by varying the parameteR $M_{\text{max}}$.
\subsubsection{Tumor Segmentation}
\begin{figure*}[!t]
	\centering
		\includegraphics[width=0.5\linewidth]{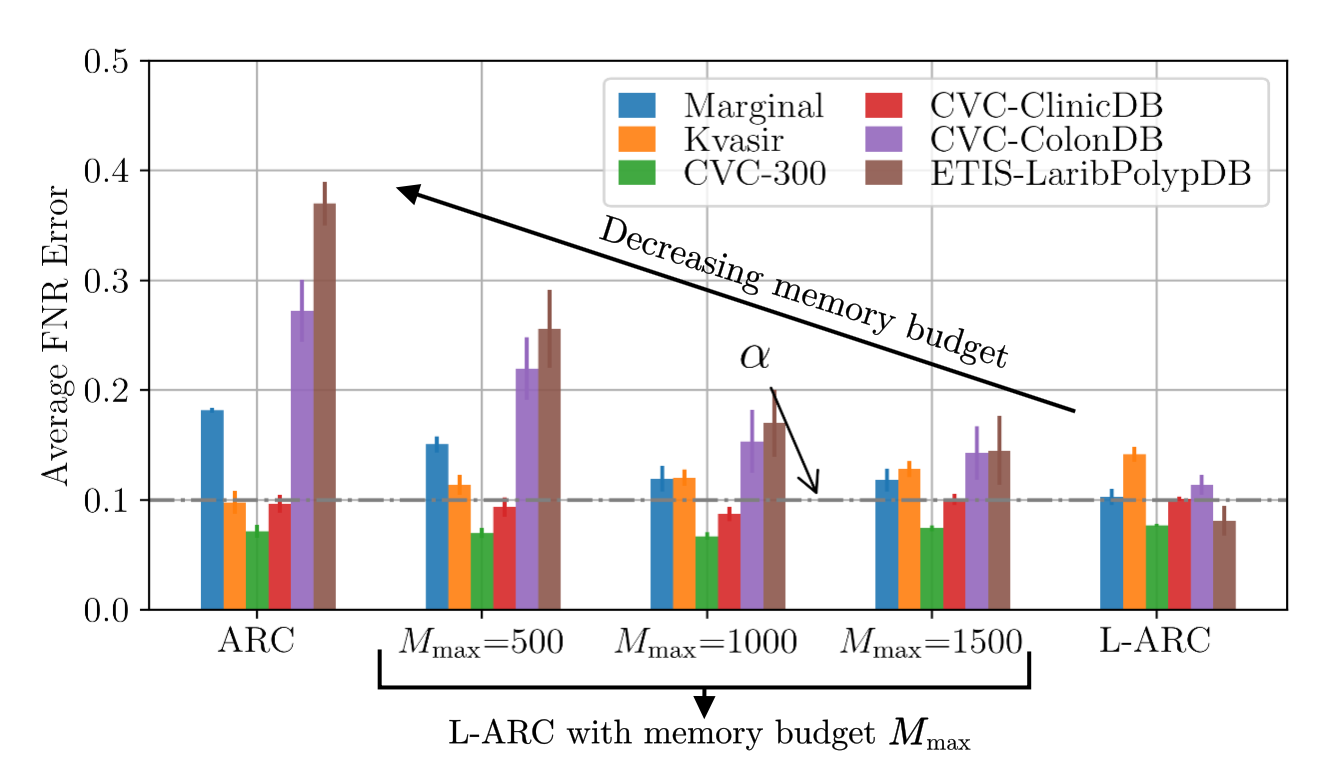}
	\caption{FNR obtained by ARC, L-ARC, and L-ARC with limited memory budget $M_{\text{max}}\in\{500,1000,1500\}$. As the memory budget increases, the localized risk control performance of L-ARC interpolates between ARC and L-ARC.}
    \label{fig:mem_eff_tumor}
\end{figure*}
Using the setup described in Section \ref{sec:exp_tumore_seg}, we now consider calibrating the image segmentation model using L-ARC with a truncated threshold (\ref{eq:threshold_truncated}). In Figure \ref{fig:mem_eff_tumor}, we report the average FNR conditioned on different data sources for $M_{\text{max}}\in\{500,1000,1500\}$. As a benchmark, we also compare against ARC and L-ARC without truncation. By adjusting the value of $M_{\text{max}}$, it is possible to trade off localized risk control for memory efficiency. In fact, the effect of truncation on L-ARC’s performance is minimal when the number of coefficients in the truncation is large $(M_{\text{max}}=1500)$. However, for greater memory savings $(M_{\text{max}}=500)$, L-ARC’s performance becomes similar to that of ARC. In all cases, L-ARC provides better localized risk control than ARC.
\subsubsection{Beam Selection}    
\begin{figure*}[!t]
	\centering
	\includegraphics[width=\linewidth]{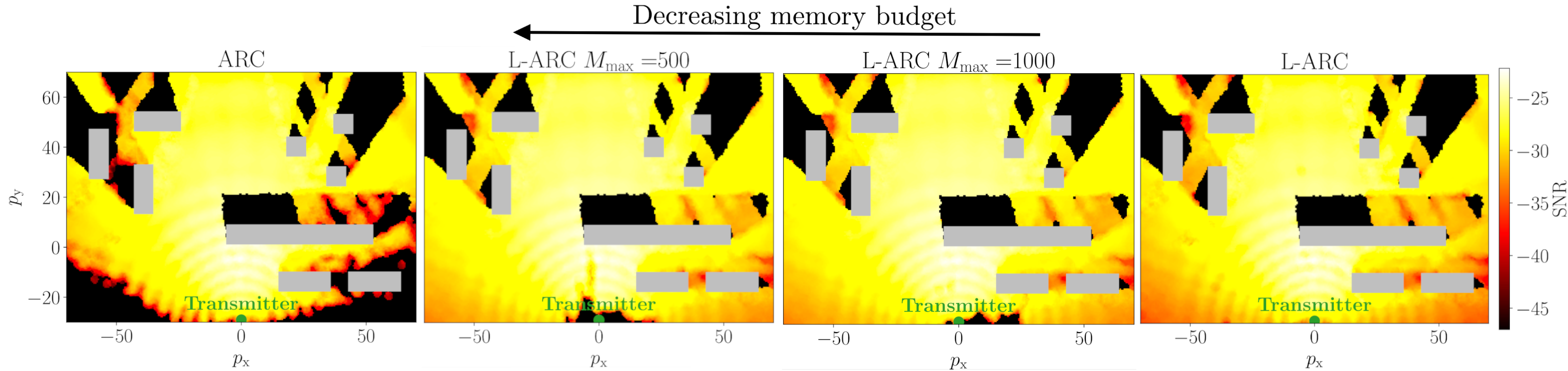}	
    \caption{SNR across the deployment attained by L-ARC with limited memory budget $M_{\text{max}}$.}
    \label{fig:mem_eff_beams}
\end{figure*}                                                                                                                                                    
We consider the beam selection problem discussed in Section \ref{sec:exp_beam_selection}. In Figure \ref{fig:mem_eff_beams}, we report the SNR levels across the deployment attained by ARC, L-ARC, and L-ARC with a truncated threshold with a maximum number of coefficients $M_{\text{max}} \in \{500, 1000\}$. As the number of coefficients $M_{\text{max}}$ and the memory requirement reduce, the localized risk control performance of L-ARC also decreases. Nonetheless, even for small $M_{\text{max}}$, L-ARC delivers a more consistent SNR level across the deployment compared to ARC.

\end{document}